\documentclass[letterpaper, 10 pt, journal, twoside]{IEEEtran}
\pdfoutput=1

\IEEEoverridecommandlockouts                              

\usepackage{amssymb,amsmath}       
\usepackage{amsthm}
\usepackage{array}
\usepackage{bm}                    
\usepackage{xcolor}
\usepackage{epsfig}
\usepackage{float}
\usepackage{graphicx,color,psfrag} 
\usepackage{multirow}              
\usepackage{tabularx}              
\usepackage{booktabs}
\usepackage{wrapfig}
\usepackage{times}

\usepackage[ruled,linesnumbered]{algorithm2e}
\let\oldnl\nl
\newcommand{\nonl}{\renewcommand{\nl}{\let\nl\oldnl}}

\usepackage[caption=false,font=footnotesize]{subfig}
\usepackage[inline]{enumitem}

\usepackage{multicol}
\usepackage[bookmarks=true]{hyperref}

\makeatletter
\def\subsubsection{\@startsection{subsubsection}
                                 {3}
                                 {\z@}
                                 {0ex plus 0.1ex minus 0.1ex}
                                 {0ex}
                                 {\normalfont\normalsize\itshape}}
\makeatother

\newcommand{\nm}[1]{\textrm{#1}}
\newcommand{\mc}[1]{\mathcal{#1}}
\newcommand{\mcs}[2]{\mathcal{#1}_\nm{#2}}

\newcommand{\mcsu}[3]{{\mathcal{#1}_\nm{#2}^{#3}}}

\newcommand{\mfs}[2]{\mathfrak{#1}_\nm{#2}}
\newcommand{\bms}[2]{\bm{#1}_\nm{#2}}
\newcommand{\bmsu}[3]{\bm{#1}_\nm{#2}^\nm{#3}}

\newcommand{\suchthat}{\,\vert\,}
\newcommand{\ie}{{\textit{i.e.}}}

\newcommand{\func}[1]{{\tt{#1}}}
\newcommand{\kw}[1]{{\tt{#1}}}
\newcounter{counter1}
\newtheorem{definition}[counter1]{Definition}
\newcounter{counter2}
\newtheorem{problem}[counter2]{Problem}
\newcommand{\comment}[1]{}

\graphicspath{{figures/}}

\renewcommand{\tt}{\fontfamily{cmtt}\selectfont}

\author{Zhou Xian, Puttichai Lertkultanon and Quang-Cuong Pham%
\thanks{Manuscript received: December, 16, 2016; Revised March, 16, 2017;
Accepted April, 14, 2017.}
\thanks{This paper was recommended for publication by Editor Nancy Amato upon
evaluation of the Associate Editor and Reviewers' comments. This work was 
supported by Tier 1 grant RG109/14 awarded by the Ministry of Education of 
Singapore.}
\thanks{This paper has supplementary downloadable material available at
http://ieeexplore.ieee.org, provided by the authors. This includes one 
multimedia MP4 format movie clip, which shows robots executing planned 
trajectories in both simulation and in real experiments. This
material is 28.5 MB in size.
}
\thanks{The authors are with the School of Mechanical and Aerospace
    Engineering, Nanyang Technological University, 50 Nanyang Avenue,
    Singapore 639798 {\tt\small zhou0202@e.ntu.edu.sg}}%
\thanks{Digital Object Identifier (DOI): see top of this page.}
}

\title{Closed-Chain Manipulation of Large Objects by Multi-Arm Robotic Systems} 

\begin{document}
\setlength{\algomargin}{2em}
\SetKwIF{If}{ElseIf}{Else}{if}{then}{elif}{else}{}
\SetKwFor{ForEach}{for each}{do}{}
\SetKw{Continue}{continue}
\SetKw{Break}{break}
\SetAlgoNoEnd
\DontPrintSemicolon

\maketitle

\begin{abstract}
  Closed kinematic chains are created whenever multiple robot arms
  concurrently manipulate a single object. The closed-chain constraint,
  when coupled with robot joint limits, dramatically changes the
  connectivity of the configuration space. We propose a regrasping move,
  termed ``IK-switch'', which allows efficiently bridging components of the
  configuration space that are otherwise mutually disconnected. This move,
  combined with several other developments, such as a method to stabilize
  the manipulated object using the environment, a new tree structure, and a
  compliant control scheme, enables us to address complex closed-chain
  manipulation tasks, such as flipping a chair frame, which is otherwise
  impossible to realize using existing multi-arm planning methods.
\end{abstract}

\begin{IEEEkeywords}
Motion and path planning, dual arm manipulation, manipulation planning
\end{IEEEkeywords}

\section{INTRODUCTION}
\label{sec:intro}
\IEEEPARstart{B}{imanual} or, more generally, multi-arm robotic systems are necessary to
manipulate large and heavy objects. It is however much more challenging
to plan and control multi-arm motions than single-arm motions, because
of the \emph{closed-chain} kinematic constraint. The closed-chain
constraint affects multi-arm motions at different levels.

First, at the ``local'' level, the feasible configurations of a
closed-chain system are restricted to a sub-manifolds of a lower
dimension than the configuration space. Thus, connecting nearby
configurations by a valid path is non-trivial, and requires projection
or differential IK
techniques~\cite{yakey2001randomized,pham2015time,mirabel2016hpp}.

Second, when using sampling-based motion planners such as
PRM~\cite{kavraki1996probabilistic} or RRT \cite{kuffner2000rrt}, one
needs to generate a large number of evenly distributed feasible
configurations (which will be connected to each other through local
motions). As the set of feasible configurations is of lower dimension,
its volume is zero, which again requires non-trivial modifications to
the sampling method. We call this the ``(connected) component'' level.

\begin{figure}
  \centering
  \subfloat[]{
    \begin{minipage}{0.09\textwidth}
      \centering
      \includegraphics[width=0.95\textwidth]{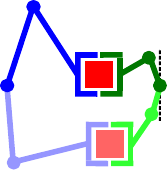}
    \end{minipage}
  }
  \subfloat[]{
    \begin{minipage}{0.17\textwidth}
      \centering
      \includegraphics[width=0.9\textwidth]{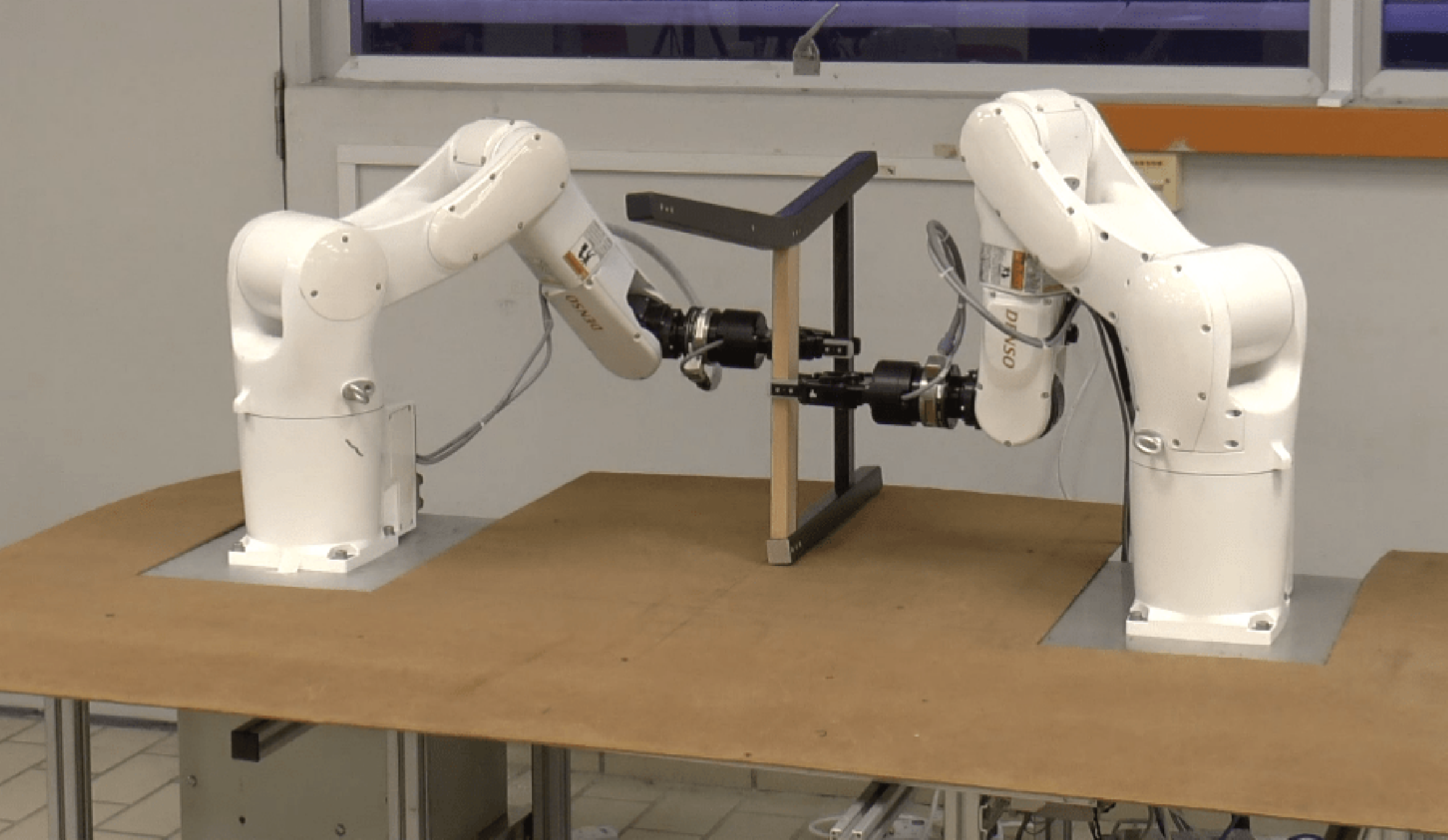}
    \end{minipage}
  }
  \subfloat[]{
    \begin{minipage}{0.17\textwidth}
      \centering
      \includegraphics[width=0.9\textwidth]{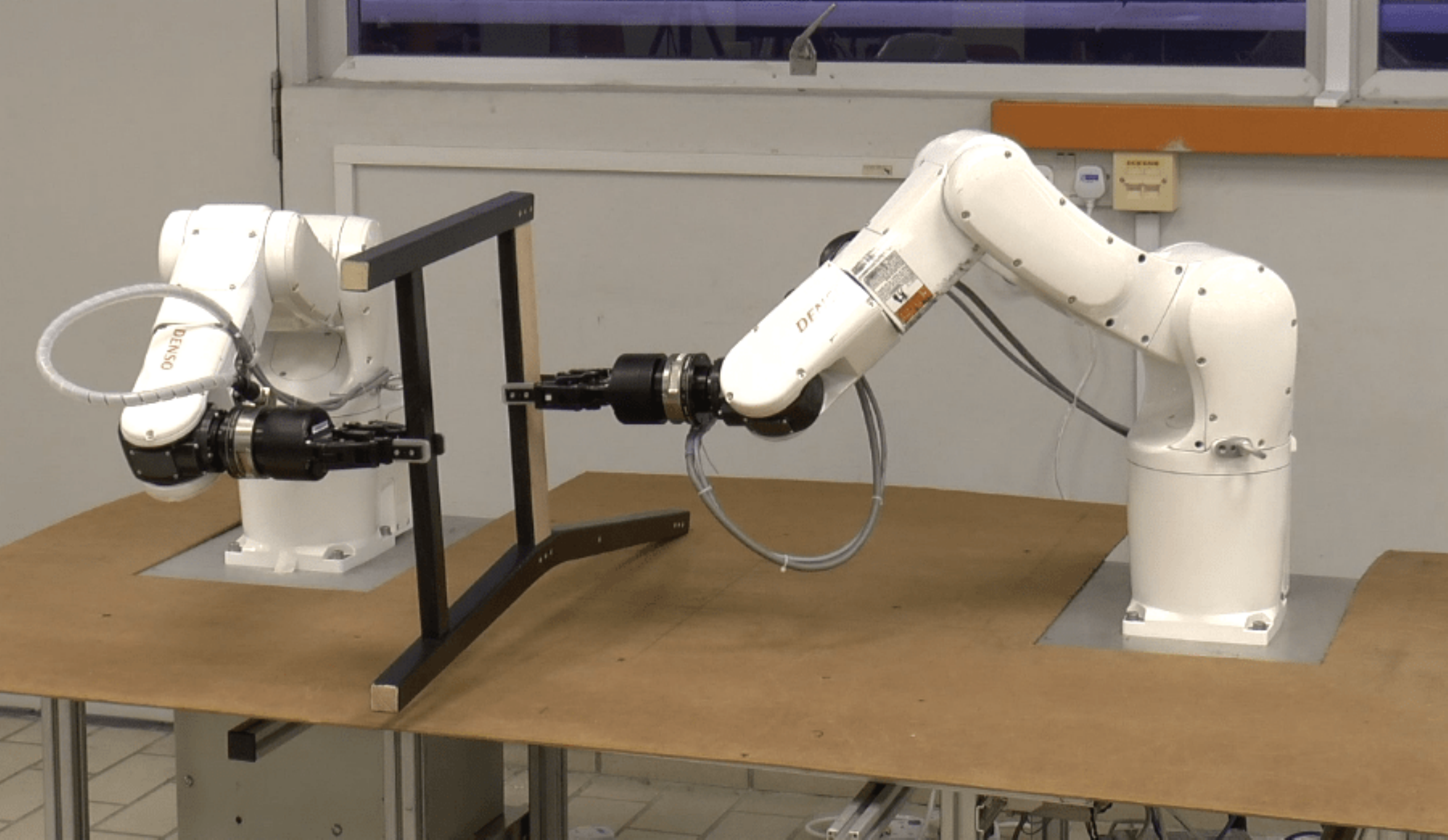}
    \end{minipage}
  }
  \caption{The closed-chain constraint, coupled with robot joint
    limits, changes the connectivity of the configuration space. (a):
    Suppose that the small green arm cannot bend backward, then the
    big blue arm cannot switch from the upper configuration to the
    lower configuration. (b,c): There is no continuous motions to flip
    a chair frame between the start configuration (b) and the goal
    configuration (c).}
  \label{fig:switch}
\end{figure}

We identify in this paper another level, termed the ``global'' level,
which encompasses different connected components that are mutually
disconnected. Indeed, the closed-chain constraint, when coupled with
robot joint limits, dramatically changes the
connectivity of the set of feasible
configurations. Fig.~\ref{fig:switch}(a) illustrates this point:
because of the closed-chain constraint and the joint limits of the
small green arm, the big blue arm cannot switch from the upper
configuration to the lower configuration.

\captionsetup[subfigure]{position=b}


With the ``local'' and ``component'' levels being relatively well
understood~\cite{han2000kinematics, yakey2001randomized,
  jaillet2013path, kim2016tangent, mirabel2016hpp}, it is the ``global'' level that
constitutes a major hurdle when deploying multi-arm systems to address
practical tasks, such as flipping a chair frame as shown in
Fig.~\ref{fig:switch}(b, c).

\subsection*{Contributions and Organization of the Paper}
\label{sec:contri}
Tasks involving large or heavy objects can be impractical for
single-arm systems because of payload and/or gripper strength
limitations. Our goal here is to develop a planning and control
framework for multi-arm systems to carry out such tasks, such as
flipping a chair frame as in Fig.~\ref{fig:switch}(b, c). For this, we
introduce the following contributions:

\begin{itemize}[leftmargin=*]
\item a move, termed ``IK-switch'', which is a regrasping move that allows
  connecting different components that are otherwise mutually
  disconnected. We argue that such ``IK-switch'' moves help address the
  ``global'' problem discussed previously;
\item a method to use the environment to help stabilize large manipulated
  objects during ``IK-switch'' moves;
\item a tree structure adapted to the ``IK-switch'' move, which allows
  accelerating the planning;
\item the integration of the above contributions into a planning and
  control framework that can tackle complex manipulations. In
  particular, we implement a compliant control scheme that allows
  executing closed-chain motions under model uncertainty. We showcase
  the framework on a difficult manipulation task: flipping a chair
  frame using two robot arms. Planning time is less than 20 seconds
  and execution is smooth, as shown in the accompanying video. We have not seen in the
  literature a demonstration of planning and execution of a bimanual
  task of such a level of complexity.
\end{itemize}

The paper is organized as follows. In Section~\ref{sec:related}, we discuss
related works in closed-chain motion planning and manipulation planning
using regrasping. In Section~\ref{sec:formulation}, we analyze in detail
how the closed-chain constraint changes the connectivity of the
configuration space and formulate the planning problem. In
Section~\ref{sec:planning}, we present the core technical contributions of
this paper that allow efficiently addressing complex closed-chain
manipulation tasks. In Section~\ref{sec:experiments}, we describe the
simulations and hardware experiments (which include the challenging task of
flipping of a chair frame using two robot arms) to validate the proposed
framework. Finally, Section~\ref{sec:conclusion} draws a conclusion of our
proposed approach.

\section{Related Works}
\label{sec:related}

{\small
\subsection{Motion Planning with the Closed-Chain Constraint}

Direct sampling in configuration space has zero probability of
generating a random configuration which satisfies closed-chain
constraints. This is due to the fact that the constraint manifold has
its dimension lower than that of the ambient
space~\cite{yakey2001randomized}. To generate a random closed-chain
configuration, the authors of~\cite{han2000kinematics} proposed to
break the closed-chain into several (open) sub-chains. A configuration
of one sub-chain can be directly sampled and the configurations of
other sub-chains computed so as to close the kinematic loop. This
method was further refined in~\cite{CSL02icra}. Random Gradient
Descent was used in~\cite{yakey2001randomized} to move a randomly
sampled configuration toward a constraint manifold. In more recent
work, \cite{jaillet2013path} and \cite{kim2016tangent} sample
configurations on a tangent space of the constraint manifold, and
\cite{mirabel2016hpp} used the Newton-Raphson method for projection on
the constraint manifold to obtain valid configurations and
paths. However, while all these planners might be able to find a path
(if it exists) within a single component, they lack the ability to
address the problem at a ``global'' level.

\subsection{Regrasping} 

Although regrasping itself is merely a robot breaking and re-initiating
contacts (grasps) with an object, how to do regrasping in such a way that
facilitates manipulation of the object into its desired goal transformation
is not trivial. Several tools, including Grasp-Placement
Table~\cite{TP87icra}, Regrasp Graph~\cite{WanX15icra}, high-level
Grasp-Placement Graph~\cite{LP15ral}, have been devised to help reason over
a large number of possible combinations of grasps and placements such that
the planner can choose only a few combinations that would sufficiently
bring the system toward the goal.

Previous work considering regraspings used such moves mainly
for the purpose of changing grasps, either one robot changing from one
grasp to another or changing from one robot grasping to another robot
grasping~\cite{SauX10iros},~\cite{HarX14icra}. In this work, we
utilize regrasping moves not necessarily to change grasps. Instead,
we use them to establish bridges between different disconnected components,
which is essentially useful in planning.

\subsection{Bimanual Manipulation Planning}

A pioneering work in this direction was published
in~\cite{koga1992experiments}. In the paper, the authors presented three
manipulation planning algorithms for two-arm robotic systems. The first
two algorithms employed exhaustive search over discretized configuration
space and therefore could only solve some simplified planar bimanual
manipulation planning problems. The third algorithm adapted the
randomized potential field technique~\cite{BL91ijrr} to work with a
closed-chain system. It used regrasping as a way to escape once trapped
in a local minimum in the potential field. Although the work itself is
interesting and the authors also provided some basic understanding and
characterization of the problem, they totally disregarded joint limits
in the planning and the planners could only cope with very limited
ranges of problems.

In~\cite{GCS08aim}, the authors presented a dual-arm motion planner
which was able to plan motions crossing different closed-chain-induced
manifolds via singular configurations. Although motions generated by
this planner will not require regrasping, the method itself relies on
the full knowldege of IK classes characterization and the ability to
sample directly singular configurations, both of which may not usually
be available in practice.

More recent work addressing bimanual manipulation planning exist. They,
however, either do not consider any closed-chain
motions~\cite{SauX10iros},~\cite{HarX14icra},~\cite{LPK15iros}, or use
heuristic search over a discretized configuration space~\cite{CCL13ijrr}
which is only capable of planning simple motions.

However, they either consider using two arms only for increasing
workspace and therefore not considering any closed-chain
motions~\cite{SauX10iros},~\cite{HarX14icra}, or use heuristic search
over a discretized configuration space~\cite{CCL13ijrr} which is only
capable of planning simple motions.
}
\section{Problem Formulation and Analysis}
\label{sec:formulation}

In this Section, we define some mathematical notations for subsequent
discussions and present a formal formulation of the problem. Moreover, we
analyze the problem by dividing it into different cases that might be
encountered in planning, and present the ``IK-switch'' move to address them.

\subsection{Closed-Chain Constraint}
\label{sec:ccc}
Consider a system consisting of $k$ robots and a movable object. Let
$\mcsu{C}{robot}{i} \subseteq \mathbb{R}^{n_i}$ be the configuration space
of the $i^\textrm{th}$ robot, where $n_i$ is its number of
degrees-of-freedom (DOF), and
$\mcs{C}{obj} \subseteq SE(3)$\footnote{$SE(3)$ denotes the special
  Euclidean group of rigid body motions in a 3-dimensional space.} the
configuration space of the object. The composite configuration space of the
system is described as
$\mcs{C}{composite} = \mcsu{C}{robot}{1} \times \mcsu{C}{robot}{2} \times
\cdots \times \mcsu{C}{robot}{k} \times \mcs{C}{obj}$. A \emph{composite
  configuration} $\bm{c} \in \mcs{C}{composite}$ can then be written as
$\bm{c} = (\bm{q}_{1},\bm{q}_{2}, \ldots, \bm{q}_{k}, \bms{T}{obj})$, where
$\bm{q}_{i} \in \mcsu{C}{robot}{i}$ is the configuration of the
$i^{\text{th}}$ robot and $\bms{T}{obj} \in SE(3)$ is the homogeneous
transformation of the object.

When all the robots are grasping the object with their end-effectors, the
system forms closed kinematic chains. In this case, the composite
configuration $\bm{c}$ implicitly determines $\bm{G}$, the set of grasping
poses of the robots. In other words, it defines the relative
transformations from the object to the end-effector of each robot. This
constraint can be described in the form $\mc{F}_{\bm{G}}(\bm{c}) = \bm{0}$, where
$\bm{0}$ is a zero vector of appropriate dimension. Let
$\mcs{C}{cc} \subset \mcs{C}{composite}$ be defined as
\begin{equation}
  \mcs{C}{cc} = \{ \bm{c} \suchthat \bm{c} \in \mcs{C}{composite},
  \mc{F}_{\bm{G}}(\bm{c}) = \bm{0}\}.
\end{equation}
Excluding singularities, $\mcs{C}{cc}$ is a set of manifolds of a lower
dimension lying in $\mcs{C}{composite}$~\cite{yakey2001randomized},
\cite{Bur89ar}.

\subsection{Essentially Mutually Disconnected (EMD) Components}

\begin{definition}
  Given a feasible configuration $\bm{c}$, we define the (connected)
  \textbf{component} $\mc{S}(\bm{c})$ as the set of all feasible
  configurations which can be reached from $\bm{c}$ by continuous and
  feasible paths (\ie, paths that are collision-free and respect the 
  closed-chain constraint and robot joint limits). Two components, $\mc{S}(\bm{c}_1)$ and $\mc{S}(\bm{c}_2)$, are \textbf{essentially mutually disconnected (EMD)} if they
  are indeed disconnected or if, in practice, one cannot find any
  connection between the two within a reasonable amount of time.
\end{definition}

Note that we use the term ``essentially'' in the above definition as it is
very difficult, in an actual problem instance, to provide a rigorous
certificate that two components are indeed disconnected. Consider the
system in Fig.~\ref{fig:switch}(a). One can clearly see that the components
containing respectively the upper configuration and the lower configuration
are disconnected. Yet a certificate of this disconnectedness would involve
complex trigonometry formulae. Such certificates are even more difficult,
if not impossible, to obtain in high-DOF systems such as in
Fig.~\ref{fig:switch}(b, c). One can however say that the configurations in
Fig.~\ref{fig:switch}(b) and Fig.~\ref{fig:switch}(c) are \emph{essentially
  mutually disconnected} after running state-of-the-art planners -- without
regrasping -- for hours without finding any solution.

In order to bridge EMD components, our planning algorithm plans not only
the component-level closed-chain motions, but also regrasping moves that
help the system ``jump'' across different EMD components. This
significantly enlarges the size of the solution space.

\subsection{Problem Formulation}

In addition to $\mcs{C}{cc}$ which satisfies the closure constraint, we
denote by $\mcs{C}{free} \subseteq \mcs{C}{composite}$ the set containing
all collision-free composite configurations. Moreover, define
$\pi: \mcs{C}{composite} \rightarrow SE(3)$ as a projection from a
composite configuration space to $\mcs{C}{obj}$ such that for
$\bm{c} = (\bm{q}_{1},\bm{q}_{2}, \ldots, \bm{q}_{k}, \bms{T}{obj})$,
$\pi(\bm{c}) = \bms{T}{obj}$. In a general closed-chain motion planning
problem for a multi-arm system, typically one is given a start composite
configuration which imposes a closure constraint on the system, and a goal
configuration of the object; with regards to the goal configuration,
grasping pose of each robot is pre-determined by the start composite
configuration, while the specific configuration is unknown In addition, in
cases when a multi-arm robot system is required, it is certain that the
object's contact stability\footnote{Contact stability here refers to the
  state that no slippage in each robot's grasping can be caused by the
  object's inertial force or gravity.} is critical. By using the notations
presented above, such a problem can be stated as follows.

\begin{problem}
  Given a start composite configuration
  $\bms{c}{start} \in \mcs{C}{cc} \cap \mcs{C}{free}$ and a goal object
  configuration $\bms{T}{obj}$, find a path
  $P: [0, 1] \rightarrow \mcs{C}{cc} \cap \mcs{C}{free}$ such that
  \begin{enumerate*}
  \item $P(0) = \bms{c}{start}$;
  \item $\pi(P(1)) = \bms{T}{obj}$; and
  \item the system maintains contact stability throughout $P$.
  \end{enumerate*}
  \label{problem:main}
\end{problem}

\subsection{Problem Analysis}

For convenience, given an EMD component $\mc{S}$, we define a projected
space $\Pi(\mc{S})$ as
$\Pi(\mc{S}) = \{ \pi(\bm{c}) \suchthat \bm{c} \in \mc{S} \}$.

Consider Problem~\ref{problem:main}. When given the goal object
transformation $\bms{T}{goal}$, there exists multiple $\bms{c}{goal}$ since
different inverse kinematic (IK) solutions exists for a certain set of
end-effector transforms. Let
$\mfs{C}{goal} = \{ \bmsu{c}{goal}{1}, \bmsu{c}{goal}{2}, \ldots,
\bmsu{c}{goal}{m} \}$ denote the collection of all $m$ possible goal
composite configurations. With regards to the relation between
$\bms{c}{goal}$ and $\mfs{C}{goal}$, two possible cases exist as follows.

\subsubsection*{Case 1}
$\exists \bms{c}{goal} \in \mfs{C}{goal}$ such that
$\bms{c}{goal} \in \mc{S}(\bms{c}{start})$.

This means that there exists a feasible path $P$ from $\bms{c}{start}$ to
some $\bms{c}{goal}$ with no regrasping. However, there might exists other
$\bm{c}'_{\text{goal}} \in \mfs{C}{goal}$ which do not lie in
$\mc{S}(\bms{c}{start})$. If the searching efficiency of a bi-directional
planner, such as a BiRRT~\cite{kuffner2000rrt}, is desired, we need to
manually select a goal configuration $\bms{c}{goal} \in \mfs{C}{goal}$
which is also in $\mc{S}(\bms{c}{start})$. However, to the best of our
knowledge, there is currently no effective method for such selection. It is
likely that for $\bms{c}{goal}$ selected based on certain heuristics,
$\bms{c}{goal} \notin \mc{S}(\bms{c}{start})$. This would fall into
sub-cases discussed in \emph{Case 2}. One possible approach to avoid
selecting goal configurations is to extend the idea of BiSpace
planning~\cite{diankov2008bispace} to such closed-chain systems. However,
this approach is limited to the condition that an ideal $\bms{c}{goal}$
exists in $\mc{S}(\bms{c}{start})$, and cannot handle the cases presented
below.

\subsubsection*{Case 2}
$\forall \bms{c}{goal} \in \mfs{C}{goal}$,
$\bms{c}{goal} \notin \mc{S}(\bms{c}{start})$.

In this case, all possible goal configurations are \emph{essentially
  mutually disconnected} from $\bms{c}{start}$. Therefore, regrasping is
necessary for the system to traverse different $\mc{S}$ to reach a goal
configuration. Consider choosing $\bms{c}{goal} \in \mfs{C}{goal}$ at
random.  There are two sub-cases as follows.

\paragraph*{Case 2.1}
$\Pi(\mc{S}(\bms{c}{start})) \cap \Pi(\mc{S}(\bms{c}{goal})) \neq
\emptyset$.

\noindent Since the intersection is not empty, there exists a path in
$\Pi(\mc{S}(\bms{c}{start})) \cup \Pi(\mc{S}(\bms{c}{goal}))$ for the
object to move from $\pi(\bms{c}{start})$ to $\pi(\bms{c}{goal})$. The
required regrasping action can be done once the object configuration is in
the intersection, with the kinematic chain jumping from
$\mc{S}(\bms{c}{start})$ to $\mc{S}(\bms{c}{goal})$.

\paragraph*{Case 2.2}
$\Pi(\mc{S}(\bms{c}{start})) \cap \Pi(\mc{S}(\bms{c}{goal})) = \emptyset$.

\noindent No path exists in
$\Pi(\mc{S}(\bms{c}{start})) \cup \Pi(\mc{S}(\bms{c}{goal}))$ to bring the
object from $\pi(\bms{c}{start})$ to $\pi(\bms{c}{goal})$.

Let $\mcsu{S}{inter}{0} = \mc{S}(\bms{c}{start})$ and
$\mcsu{S}{inter}{p+1} = \mc{S}(\bms{c}{goal})$. The problem is solvable if
and only if there exists $p \geq 1$ intermediate EMD components, denoted by
$\mcsu{S}{inter}{1}, \mcsu{S}{inter}{2}, \ldots, \mcsu{S}{inter}{p}$, such
that
\begin{equation}
  \Pi(\mcsu{S}{inter}{i}) \cap \Pi(\mcsu{S}{inter}{i + 1}) \neq \emptyset
  \quad \forall i \in \{ 0, 1, \ldots, p \}.
\end{equation}
With the aid of theses intermediate EMD components, an object path can be
found in $\bigcup_{i = 0}^{p + 1} \Pi(\mcsu{S}{inter}{i})$. The
intersections between projections of these components provide shared
regions to bridge themselves together and in turn, connects
$\mc{S}(\bms{c}{start})$ and $\mc{S}(\bms{c}{goal})$.

The discussion above summarizes possible scenarios that may be encountered
in a planning problem. Note that in order to maintain contact stability,
our planner will try to find feasible placement configurations for the
object to seek support from the environment whenever a regrasping is
necessary. Consider the case where a regrasping is needed to connect
$\mc{S}(\bms{c}{1})$ and $\mc{S}(\bms{c}{2})$, the regrasping should be
performed at a composite configuration $\bm{c}$ such that $\pi(\bm{c})$ is
a valid placement configuration, where regrasping trajectory can be found
and the object remains in static equilibrium throughout the regrasping
process.

\subsection{IK-switch}
\label{sec:ikswitch}

\begin{figure}[h]
\centering
\includegraphics[width=0.3\textwidth]{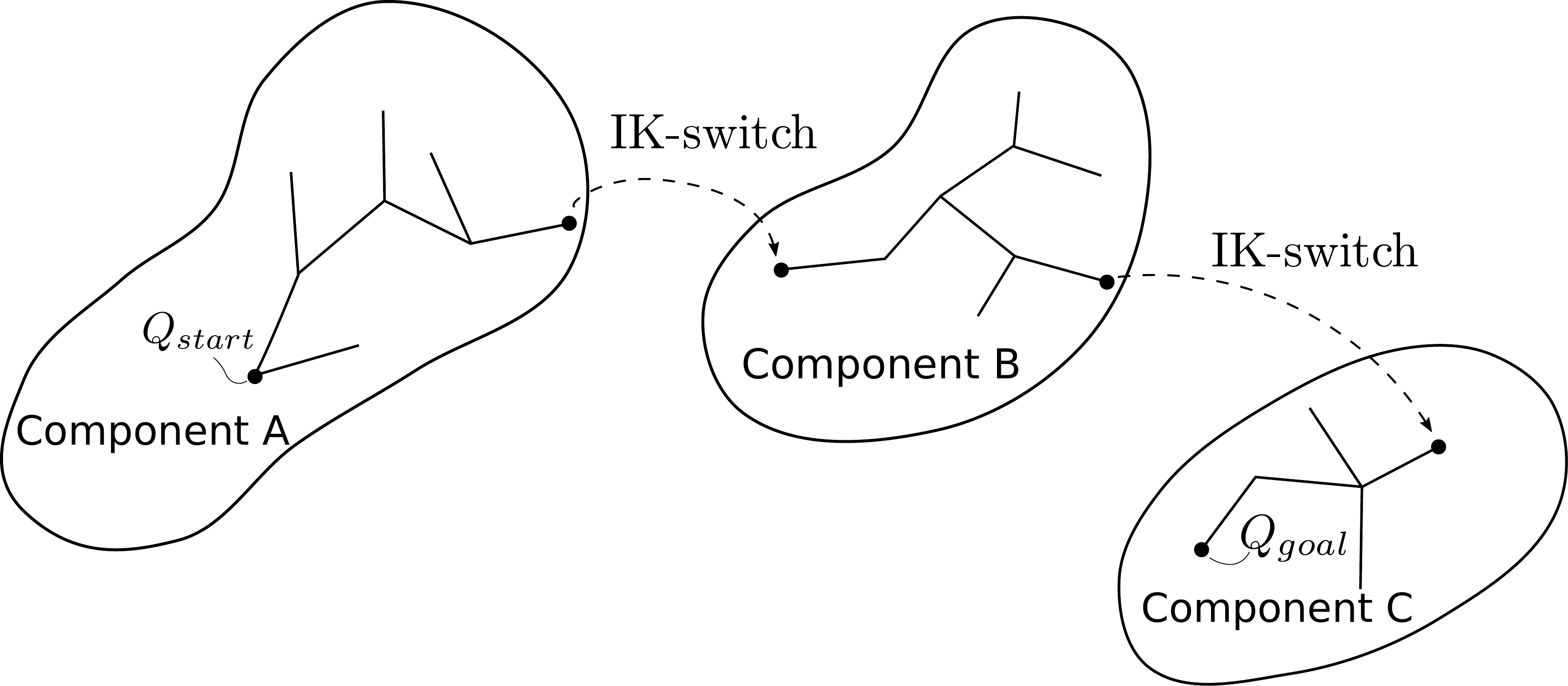}
\caption{Bridging different EMD components using IK-switch moves.}
\label{fig:IK-switch}
\vspace{-8pt}
\end{figure}

We now discuss in detail the ``IK-switch'' move that allows bridging
different EMD components, see Fig.~\ref{fig:IK-switch}. For a given 6D
end-effector pose (translation and rotation), a robot arm with 6 revolute
joints has up to 16 IK solutions. Different IK solutions can belong to
different EMD components, see e.g. Fig.~\ref{fig:switch}(a). Therefore,
``jumping'' between different IK solutions corresponding to the same
end-effector pose can bridge different EMD components.

Specifically, an ``IK-switch'' move consists in: (i) one of the
robot arm, say manipulator A, releases the grasp; (ii) manipulator A
moves freely from the original IK solution to another IK solution,
corresponding to the same grasping pose; (iii) manipulator A regrasps
the object with the same initial grasping pose.
Section~\ref{sec:IK-switch} details the implementation.

The main difference with usual regrasping moves~\cite{TP87icra, LP15ral} is
that ``IK-switch'' regrasps exactly using the same end-effector grasping
pose. This has two advantages. First, one avoids the computational
explosion of the grasp-placement table when the number of grasp classes
increases while staying expressive enough to solve difficult tasks, as
shown in Section~\ref{sec:experiments}. Second, this strategy can be used
even when one has no information on the grasp structure of the manipulated
object (e.g. grasp database, grasp classes): one single grasping pose is
used per manipulator during the whole manipulation.

Note that in step (i) of the move, as one manipulator releases the grasp,
contact with the environment is needed for the object to stay in static
equilibrium to ensure contact
stability. Section~\ref{sec:staticequilibrium} details the implementation.

Finally, one can note that the concept of jumping between different IK
solutions bears some resemblance with the transition between different
self-motion manifolds~\cite{Bur89ar} through
singularities~\cite{Tho93ck}.

\section{Path Planning} 
\label{sec:planning}

Based on the previous analysis, we propose here a planner that can address
complex multi-arm manipulation tasks. Our planner is derived from the
classical BiRRT structure and comprises two planning stages. In the first
stage, it plans a global path for the closed-chain system, which includes
\begin{enumerate*}
\item segments storing closed-chain motions without breaking the chain and
\item the vertices connecting these segments, including necessary IK-switch
  regrasping requests.
\end{enumerate*} 
In the second stage, it completes the global path by planning IK-switch
moves at tree vertices where regrasping is needed. Delaying IK-switch
planning to the second stage helps improve efficiency of the planner
sharply since most of regrasping requests will not be in the final path
connecting $\bms{c}{start}$ and $\bms{c}{goal}$. Note that planning for
IK-switch at these connecting vertices might fail when no feasible
regrasping moves can be found. In this case, we implement an efficient data
structure (see Section~\ref{sec:newtreestructure}) that re-organizes all
vertices stored in itself to retain the space information obtained
previously before the planner returns to the first stage and re-plans a new
global path.

\newcommand{\Rmax}{R_{\nm{max}}}
\newcommand{\Tf}{\mc{T}_f}
\newcommand{\Tb}{\mc{T}_b}
\vspace{-6pt}
\begin{algorithm}
  \footnotesize
  \caption{Global Path Planner}
  \label{algo:global}
  \Indm
  {\nonl{\func{Plan}($\bms{c}{start}, \bms{c}{goal}, N_{\nm{max}}, \Rmax, E$)}}\;
  \Indp

  \func{Init}($\Tf, \bms{c}{start}$), \func{Init}($\Tb, \bms{c}{goal}$)\;  
  \For{\upshape $i \gets 1$ \textbf{to} $N_{\nm{max}}$}
  {
    $\bms{T}{rand} \gets$ \func{SampleSE3Config()}\;
    $\mcs{V}{new} \gets$ \func{Extend}($\Tf, \bms{T}{rand}, \Rmax$)\;
    \If{\upshape $\mcs{V}{new}$ \textbf{is not} \kw{None}}
    {
      $\mcs{P}{connect} \gets$ \func{Connect}($\mcs{V}{new}, \Tb, \Rmax$)\;
      \If{\upshape $\mcs{P}{connect}$ \textbf{is not} \kw{None}}
      {
        \If{\upshape \func{PlanIKSwitch}($\Tf, \Tb, E, \mcs{V}{fail}$)}
        {
          \Return{\upshape \func{GeneratePath}($\Tf, \Tb, \mcs{P}{connect}$)}\;
        }
        \Else
        {
          \func{Reorganize}($\Tf, \Tb, \mcs{P}{connect}$)
        }
      }
    }
    Swap($\Tf$, $\Tb$)\;
  }
  \Return{\upshape \kw{Failure}}\;

  \vspace{2pt}
  \setcounter{AlgoLine}{0}
  \Indm
  {\nonl{\func{Extend}($\mc{T}_{f}, \bms{T}{rand}, R_{\nm{max}}$)}}\;
  \Indp
  
  $\mcs{V}{near} \gets$ \func{NearestNeighbor}($\Tf, \bms{T}{rand}, \Rmax$)\;
  $\bms{T}{goal} \gets$
  \func{ComputeGoalSE3Config}($\pi(\mcs{V}{near}.\bm{c}), \bms{T}{rand}$)\;

  \If{\upshape $\bms{T}{goal}$ \textbf{is not} \kw{None}}{
    $\mcs{P}{SE3} \gets$ \func{InterpolateSE3Path}($\pi(\mcs{V}{near}.\bm{c}), \bms{T}{goal}$)\;
    
    \If{\upshape $\mcs{P}{SE3}$ \textbf{is not} \kw{None}}{

      $status \gets$ \func{ComputePath}($\mcs{V}{near}.\bm{c}, \mcs{P}{SE3},
      \mcs{P}{composite}, \bms{c}{regrasp}$)\;
  
      \If{\upshape $status ==$ \kw{REACHED}}
      {
        $\mcs{V}{new} \gets$ \func{Vertex}($\bms{T}{goal}, \mcs{P}{composite}$)\;
        $\Tf.$\func{AddVertex}($\mcs{V}{new}$)\;
        \Return{$\mcs{V}{new}$}\;
      }
      \ElseIf{\upshape $status ==$ \kw{NEED\_REGRASP} \textbf{and}
        $\mcs{V}{near}.$\func{RegraspCount} $< \Rmax$}
      {
        $\mcs{V}{new} \gets$ \func{Vertex}($\pi(\bms{c}{regrasp}), \mcs{P}{composite}$)\;
        $\mcs{V}{new}.$\kw{NeedRegrasp} $\gets$ \kw{True}\;
        $\mcs{V}{new}.$\kw{RegraspCount} $\gets$
        $\mcs{V}{new}$.\kw{Parent}.\kw{RegraspCount}$ + 1$\;
        $\Tf.$\func{AddVertex}($\mcs{V}{new}$)\;
        \Return{$\mcs{V}{new}$}\;
      }
    }
  }
  \Return{\upshape \kw{None}}\;

  \vspace{2pt}
  \setcounter{AlgoLine}{0}
  \Indm
  {\nonl{\func{Connect}($\mcs{V}{new}, \Tb, R_{\nm{max}}$)}}\;
  \Indp
  
  $\mcs{V}{near} \gets$ \func{NearestNeighbor}($\Tb,
  \pi(\mcs{V}{new}.\bm{c}), \Rmax - \mcs{V}{new}$.\kw{RegraspCount})\;
  
  $\mcs{P}{SE3} \gets$ \func{InterpolateSE3Path}($\pi(\mcs{V}{near}.\bm{c}),
  \pi(\mcs{V}{new}.\bm{c})$)\;
  
  \If{\upshape $\mcs{P}{SE3}$ \textbf{is not} \kw{None}}
  {
    $status \gets$ \func{ComputePath}($\mcs{V}{near}.\bm{c}, \mcs{P}{SE3},
    \mcs{P}{connect}, \bms{c}{regrasp}$)\;
    
    \If{\upshape $status ==$ \kw{REACHED}}
    {
      \If{\upshape $\mcs{P}{connect}.\bms{c}{end} == \mcs{V}{new}.\bm{c}$}
      {
        \Return{\upshape $\mcs{P}{connect}$}\;
      }
      \ElseIf{\upshape $\mcs{V}{new}$.\kw{RegraspCount} +
        $\mcs{V}{near}$.\kw{RegraspCount} $< R_{\nm{max}}$}
      {
        $\mcs{P}{connect}$.\kw{NeedRegrasp} $\gets$ \kw{True}\;
        \Return{\upshape $\mcs{P}{connect}$}\;
      }
    }
  }
  \Return{\upshape \kw{None}}\;
\end{algorithm}

\subsection{Stage 1: Planning Global Path}
\label{sec:globalpath}

Given a planning query determined by $\bms{c}{start}$ and $\bms{T}{goal}$,
we first pick one composite configuration $\bms{c}{goal}$ such that
$\pi(\bms{c}{goal}) = \bms{T}{goal}$, either randomly or according to
certain heuristics, such as picking the one closest to $\bms{c}{start}$. In
a multi-robot system described previously, the movable object is the nexus
linking all the robots. Thus, we take a decomposed approach to plan
closed-chain motions: the planner plans a rigid body motion of the object
in $SE(3)$ first, then enforces the robots to follow the object's path. The
flow of the global path planner is summarized in Algorithm
\ref{algo:global}.

This global path planner grows two trees rooted at $\bms{c}{start}$ and
$\bms{c}{goal}$. $N_{\nm{max}}$ is the maximum number of iterations allowed
for tree extension. The planner also takes as its input a parameter
$\Rmax$, which sets the maximum number of regraspings allowed\footnote{In
  our planner, one regrasping is defined as a single jump from one
  component $\mc{S}$ to another, therefore possibly containing multiple
  robots performing regrasping at a single object
  configuration.}. Parameter $E$ is a description of the workspace
environment, which will be used for planning IK-switch moves later. Some
key functions in planning for global path are explained below.

{\small

  \begin{itemize}[leftmargin=*]
  \item \func{SampleSE3Config} samples a transformation matrix in
    $SE(3)$. We separately sample orientation and translation parts. The
    orientation is uniformly sampled from Special Orthogonal Group
    $SO(3)$~\cite{kuffner2004effective} while the translation is uniformly
    sampled from a user-defined range.

  \item \func{NearestNeighbor} searches over all vertices in the given tree
    and returns the one with a transformation closest to
    $\bms{T}{rand}$. We use the distance metric which is a combination of
    Euclidean distance (for translation part) and the minimal geodesic
    distance in $SO(3)$ (for rotation part; see~\cite{PR97acmtg} for more
    details). To limit total number of regraspings in the final path, this
    function also takes $\Rmax$ as an input and ignores any vertices having
    \kw{RegraspCount} greater than $\Rmax$.

  \item \func{InterpolateSE3Path} generates an $SE(3)$ trajectory
    connecting the given transformations. The procedure used here is
    similar to the one in~\cite{nguyen2016time}.

  \item \func{ComputePath} \label{func:cppath}
    generates a composite path required for the
    motion of the closed-chain system. The detail implementation is listed
    in Algorithm~\ref{algo:computepath}. It discretizes the input object
    path into a series of transformations, according to a pre-defined time
    step, and stores them in a list $\bm{L}_{\bm{T}}$. It iterates through
    $\bm{L}_{\bm{T}}$ and for each $\bms{T}{obj}$, it calls
    \func{ComputeCompositeConfig} to compute corresponding composite
    configurations. At each time instant, we use a differential IK
    algorithm~\cite{SicX09book,pham2015time} to generate a new IK solution for each
    robot.

    \noindent We use a differential IK solver to ensure that each newly
    generated composite configuration remains in the same $\mc{S}$ as the
    previous ones. When the solver fails, we use \func{IsNearBoundary} to
    check if the failure is because some robots reach their configuration
    space boundaries. If this is the case, \func{GetRegraspConfig} will
    compute the most flexible composite configuration
    $\bms{c}{regrasp}$. In particular, it computes all IK solutions (via
    OpenRAVE IKFast~\cite{diankov2008openrave}) for the robot $index$ to
    grasp the object at $\pi(\bm{c})$. Then in \func{SelectMostFlexibleIK},
    we use a scoring function as a heuristic to choose the best solution
    (the one with highest score). Given the lower and upper joint limits of
    the robots, $\bms{q}{l}$ and $\bms{q}{u}$, the score for a
    configuration $\bm{q}$ is
    $(\bms{q}{u} - \bm{q})^{\top}(\bm{q} - \bms{q}{l})$. Then a composite
    path $\mcs{P}{composite}$ can be generated from all feasible
    configurations in $\bm{L}_{\bm{c}}$ together with a IK-switch request.

  \item \func{Connect}: When the planner attempts to connect the backward
    tree $\Tb$ to a given vertex in $\Tf$, it computes an allowed number of
    regraspings first for \func{NearestNeighbor} to select a vertex from
    $\Tb$. Similar to \func{Extend}, it uses the differential IK solver to
    compute a composite path $\mcs{P}{connect}$ from $\mcs{V}{near}$ to
    $\mcs{V}{new}$. If discrepancy exists between the last configuration in
    $\mcs{P}{connect}$ and the one stored in $\mcs{V}{new}$, we add one
    more regrasping request to $\mcs{P}{connect}$ if the limit is not
    exceeded.    

  \end{itemize}
  
}

After $\mcs{P}{connect}$ is returned from \func{Connect}, a global path is
considered found, then the planner enters the second planning stage to plan
for IK-switch moves.

\vspace{-6pt}
\begin{algorithm}
  \footnotesize
  \caption{Composite Path Computation}
  \label{algo:computepath}
  \Indm
  {\nonl{\func{ComputePath}($\bms{c}{start}, \mcs{P}{SE3}, \mcs{P}{composite}, \bms{c}{regrasp}$)}}\;
  \Indp
  
  $\bm{L}_{\bm{T}}\gets$ \func{Discretize}($\mcs{P}{SE3}$)\;
  $\bm{L}_{\bm{c}} \gets$ \func{EmptyList}()\;
  $\bms{c}{prev} \gets \bms{c}{start}$\;
  \ForEach{\upshape $\bms{T}{obj}$ \textbf{in} $\bm{L}_{\bm{T}}$}
  {
    $\bms{c}{next} \gets$ \func{ComputeCompositeConfig}($\bms{c}{prev}, \bms{T}{obj}$)\;
    \If{\upshape $\bms{c}{next}$ \textbf{is not} \kw{None}}
    {
      $\bm{L}_{\bm{c}}.$\func{Append}($\bms{c}{next}$) \;
    }
    \ElseIf{\upshape \func{IsNearBoundary}($\bms{c}{prev}, index$)}
    {
      $\bms{c}{regrasp} \gets$ \func{GetRegraspConfig}($\bms{c}{prev}, index$)\;
      \If{\upshape $\bms{c}{regrasp}$ \textbf{is not} \kw{None}}
      {
        $\mcs{P}{composite} \gets$ \func{CompositePath}($\bm{L}_{\bm{c}}$)\;
        \Return{\upshape \kw{NEED\_REGRASP}}\;
      }
      \Else
      {
        \Return{\upshape \kw{TRAPPED}}\;
      }
    }
    \Else
    {
      \Return{\upshape \kw{TRAPPED}}\;
    }
    $\bms{c}{prev} \gets \bms{c}{next}$\;
  }
  $\mcs{P}{composite} \gets$ \func{CompositePath}($\bm{L}_{\bm{c}}$)\;
  \Return{\upshape \kw{REACHED}}\;
  
  \vspace{2pt}
  \setcounter{AlgoLine}{0}
  \Indm
  {\nonl{\func{GetRegraspConfig}($\bm{c}, index$)}}\;
  \Indp
  $\bms{L}{IK} \gets$ \func{ComputeIKs}($\pi(\bm{c}), index$)\;
  \If{\upshape $\bms{L}{IK}$ \textbf{is empty}}
  {
    \Return{\upshape \kw{None}}
  }
  $\bm{q}_{i} \gets$ \func{SelectMostFlexibleIK}($\bms{L}{IK}$)\;
  $\bms{c}{regrasp} \gets$ \func{UpdateCompositeConfig}($\bm{c}, \bm{q}_{i}, index$)\;
  \Return{\upshape $\bms{c}{regrasp}$}\;
\end{algorithm}

\subsection{Stage 2: Planning IK-Switch Moves}
\label{sec:IK-switch}

\subsubsection{Path Planning}
The planner calls \func{PlanIKSwitch} to generate paths for all the
IK-switch requests in the global path computed in the first
stage. Algorithm~\ref{algo:local} presents the detailed flow. Key functions
are discussed below.

{\small

  \begin{itemize}[leftmargin=*]
  \item \func{GlobalPathVertices} extracts the list of all vertices along
    the path connecting $\Tf$ and $\Tb$.

  \item \func{SamplePlacementConfig}: Given an object configuration
    $\bms{T}{obj}$, this function computes a valid placement configuration
    $\bms{T}{place}$ which is supposedly close to $\bms{T}{obj}$. We
    describe the procedure in more detail in
    Section~\ref{sec:staticequilibrium}.

  \item \func{ComputePath2} This function computes both the composite path
    moving from the pre-regrasping configuration to the placement
    configuration, and the path moving back to the post-regrasping
    configuration.

  \item \func{PlanRegraspPath} plans regrasp paths for robots using a vanilla
    BiRRT planner.
  \end{itemize}
  
}





The IK-switch planner returns \kw{True} if regrasping requests stored in
all vertices along the global path are found. Otherwise, it returns
\kw{False} and stores the failed vertex into $\mcs{V}{fail}$.

\begin{algorithm}
  \footnotesize
  \caption{IK-Switch Planner}
  \label{algo:local}
  \Indm
  {\nonl{\func{PlanIKSwitch}($\Tf, \Tb, E, \mcs{V}{fail}$)}}\;
  \Indp
  \ForEach{\upshape $\mc{V}$ \textbf{in} \func{GlobalPathVertices}($\Tf, \Tb$)}
  {
    \If{\upshape $\mc{V}.$\kw{NeedRegrasp} \textbf{and not} $\mc{V}.$\kw{HasRegrasp}}
    {
      $success \gets$ \kw{False}\;
      \For{\upshape $i \gets 1$ \textbf{to} $N_{\nm{max}}$}
      {
        $\bms{T}{place} \gets$ \func{SamplePlacementConfig}($\pi(\mc{V}.\bm{c}), E$)\;
        \If{\upshape $\bms{T}{place}$ \textbf{is} \kw{None}}
        {
          $\mcs{V}{fail} \gets \mc{V}$\;
          \Return{\upshape \kw{False}}\;
        }

        \{$status, \mcs{P}{go}, \mcs{P}{back}$\} =
        \func{ComputePath2}(\upshape $\mc{V}.\bm{c}, \bms{T}{place}, \mc{V}$.\kw{Child}.$\bm{c}$)\;

        \If{\upshape $status =$ \kw{failure}}
        {
          \Continue\;
        }
        
        $\mcs{P}{regrasp} \gets$ \func{PlanRegraspPath}($\mcs{P}{go}, \mcs{P}{back}$)\;
        \If{\upshape $\mcs{P}{regrasp}$ \textbf{is} \kw{None}}
        {
          \Continue\;
        }
        
        $\mc{V}.$\kw{HasRegrasp} $\gets$ \kw{True}\;
        $\mc{V}.$\func{AddRegraspAction}($\mcs{P}{go}, \mcs{P}{regrasp}, \mcs{P}{back}$)\;
        $success \gets$ \kw{True}\;
        \Break\;
      }
      \If{\upshape \textbf{not} $success$}
      {
        $\mcs{V}{fail} \gets \mc{V}$\;
        \Return{\upshape \kw{False}}\;
      }
    }
  }
  \Return{\upshape \kw{True}}\;
\end{algorithm}

\subsubsection{Placement Configuration Computation \& Static Equilibrium Checking}
\label{sec:staticequilibrium}

The function \func{SamplePlacementConfig} proceeds in iterations. In each
iteration, it injects some small random perturbation to $\bm{T}$ to get
$\bms{T}{obj}$ (except for the first iteration where it does nothing). The
function then computes a close placement configuration $\bms{T}{place}$ and
checks if $\bms{T}{place}$ is feasible (\ie, is reachable by the robots and
is in static equilibrium, as discussed below). If $\bms{T}{place}$ is
feasible, it is returned. Otherwise, it continues until some maximum number
of iterations is reached.

When performing regrasping at $\bms{T}{place}$, there can be at least one
robot holding the object at any time instant. Therefore, $\bms{T}{place}$
does not necessarily have to be \emph{stable}\footnote{A stable object
  placement is an object configuration in which the projection of the
  center of mass (COM) of the object (onto the supporting surface) lies
  inside the supporting area of the object.}. Instead, it only needs to
stay in static equilibrium with the help of contact forces provided by
grasping robots. Considering this, we can explore all types of contact with
the convex hull of the object and the supporting surface (e.g. a floor or a
table): face-face, edge-face, and vertex-face. For example, if we consider
a contact of type edge-face, we proceed by finding the edge $e$ of the
convex hull of the object at $\bms{T}{obj}$ closest to the supporting
surface $S$. Then it adds small rotation to the object such that $e$ is
parallel to $S$. Then the object is translated until it touches the
surface. The resulting transformation is $\bms{T}{place}$. In our
implementation, we compute several such transformations by adding small
translational perturbations, parallel to $S$, to the original
$\bms{T}{place}$. Then we check each one of them for static equilibrium.

To check static equilibrium of a placement configuration $\bms{T}{place}$
given that some robots are grasping the object, we reformulate Newton-Euler
equations as a matrix equation
\begin{equation}
  \label{eq:wgi}
  \underbrace{
    \begin{bmatrix}
      -m\bm{g}\\
      -m(\bm{p}_{\text{COM}} \times \bm{g})
    \end{bmatrix}
  }_{\bms{w}{GI}} =
  \underbrace{
    \begin{bmatrix}
      \bm{I}_3 & \bm{I}_3 & \cdots & \bm{I}_3\\
      [\bm{p}_1] & [\bm{p}_2] & \cdots & [\bm{p}_k]
    \end{bmatrix}
  }_{\bm{G}}
  \bms{f}{all},
\end{equation}
where $m$ is the mass of the object, $\bm{g}$ the acceleration due to
gravity, $\bm{p}_{\text{COM}}$ the position of the COM of the object at
$\bms{T}{place}$, $\bm{p}_{i}$ the position of the $i^{\text{th}}$ contact
point, $\bm{f}_{i}$ the force exerted on the object at $\bm{p}_{i}$, $k$
the total number of contact points, $\bms{w}{GI}$ is the gravito-inertial
wrench~\cite{CPN15rss},
$\bms{f}{all} = [\bm{f}_1^\top \, \bm{f}_2^\top \, \cdots \,
\bm{f}_k^\top]^\top$, and the operator $[\cdot]$ maps a vector in
$\mathbb{R}^3$ to a $3 \times 3$ skew-symmetric matrix. Note that we can
take into account surface contacts by considering forces at the vertices of
the contact area~\cite{CPN15rss}.

All the constraints related to contact forces (linearized friction-cone
constraints and the max grip force constraint) can also be written as a
vector inequality constraint $\bm{U}\bms{f}{all} \preceq \bm{b}$. Then the
configuration in consideration is in static equilibrium if there exists
$\bms{f}{all}$ satisfying \ref{eq:wgi} and all constraints. This
feasibility problem can be solved via a linear programming solver or other
available methods (c.f.~\cite{CPN15rss, PTM16icra}).

\subsection{Handling Failures in IK-Switch Planning}
\label{sec:newtreestructure}
{\small
After a global path is found, the IK-switch planner starts planning
IK-switch moves along the global path, from $\mcs{V}{start}$ in the forward
tree $\Tf$ to $\mcs{V}{goal}$ in the backward tree $\Tb$. In cases where a
failure is encountered in this planning stage, the planner will return back
to the first stage to find a new global path. In such cases, the failed
vertex $\mcs{V}{fail}$, of which the requested IK-switch cannot be found is
useless; furthermore, all the child vertices in the subtree rooted at
$\mcs{V}{fail}$ become disconnected to $\Tf$. In order to handle such
failure properly so that all the information stored in these vertices can
be retained for future use in regenerating a global path, we adopt a new
variation of the BiRRT structure. In particular, our tree re-organizes
itself in case of failure. It abandons only the edge connecting
$\mcs{V}{fail}$ and its parent vertex, and keeps the edge connecting
the two trees. This way, the disconnected subtree becomes
a subtree of $\Tb$. These two new $\Tf$ and $\Tb$ will then be used for
re-planning the global path. A blacklisted region is set within a certain
radius from $\pi(\mcs{V}{fail}.\bm{c})$; no IK-switch request will be
allowed in this region in subsequent planning. If all IK-switch requests
are solved in $\Tf$, the IK-switch planner starts from $\mcs{V}{goal}$ and
deals with $\Tb$ similarly. This data structure helps improve our planner's
efficiency substantially, since with all the vertices retained, the planner
only needs to find a path to bypass $\mcs{V}{fail}$ so as to find a new
global path. This idea would also be applicable to all other similar
planning strategies containing multiple planning stages.
}
\subsection{Properties of the Planner}
\label{sec:properties}
{\small
The complexity of the planning problem is dependant on multiple properties of
its configuration space, including dimension, expansiveness~\cite{hsu1997path}
and e-goodness~\cite{latombe1998randomized}. With regards to the dimension, 
the complexity depends exponentially on the dimension of the configuration 
space of the robot system~\cite{lavalle2006planning}. Increasing the system's 
DOF will therefore exponentially increase the complexity, hence the planning 
time.

At the local and connected-component levels, our planner relies on
variants of the classical RRT algorithm. It therefore inherits, at
these levels, many of RRT's key properties, such as spatial bias towards 
unexplored regions. Our current implementation does not guarantee 
probabilistic completeness as we sample configurations in the object space 
while growing trees in the composite configuration space; however, other 
complete variants can be used at a cost of increase in computational time. At 
the global level, to obtain completeness guarantees might require considering 
grasps classes and enumerating all possible IK solutions during switches.

We introduced a simple heuristic in Section~\ref{sec:globalpath} for
selecting an IK solution for regrasping. Selecting a ``bad'' IK (one
that does not come with enough space for the transfer motion) might
result in more frequent IK-switch moves, which in turn will adversely
affect the planning speed as well as the quality of the planned
trajectory.
}

\section{Experiments}
\label{sec:experiments}
In this Section, we illustrate the effectiveness of our planner by
solving two difficult bimanual tasks. The planner was implemented in
Python. The open-source code is available at
\url{https://github.com/quangounet/bimanual}.

We used OpenRAVE~\cite{diankov2008openrave} environment as a
test-bed. All simulations were run on a desktop computer with a $4.0$
GHz Intel\textsuperscript{\circledR} Core{\texttrademark} CPU.

In addition, we present a compliant control strategy for closed-chain
motion execution, together with hardware demonstration, in
Section~\ref{sec:control}.

\subsection{Experimental Setup}
\label{sec:expsetup}
{\small

Our bimanual robotic platform consists of two 6-DOF industrial manipulators
Denso VS060. Each manipulator is equipped with a Robotiq 2-Finger 85
Gripper, with a gripping force ranging from 30 to 100 N. One ATI Gamma
Force-Torque (F/T) sensor is attached between the end of each manipulator
and its gripper.

The distance between the two robots is optimized to be $d=1.042$ m, in
order to maximize each robot's manipulability and the bimanual system's
reachability, following procedures presented in a related work
\cite{su2016framework}.
}

\subsection{Task Description and Planning Results}
\label{sec:task}
\begin{figure}
\centering
  \subfloat[]{\includegraphics[width=0.12\textwidth]{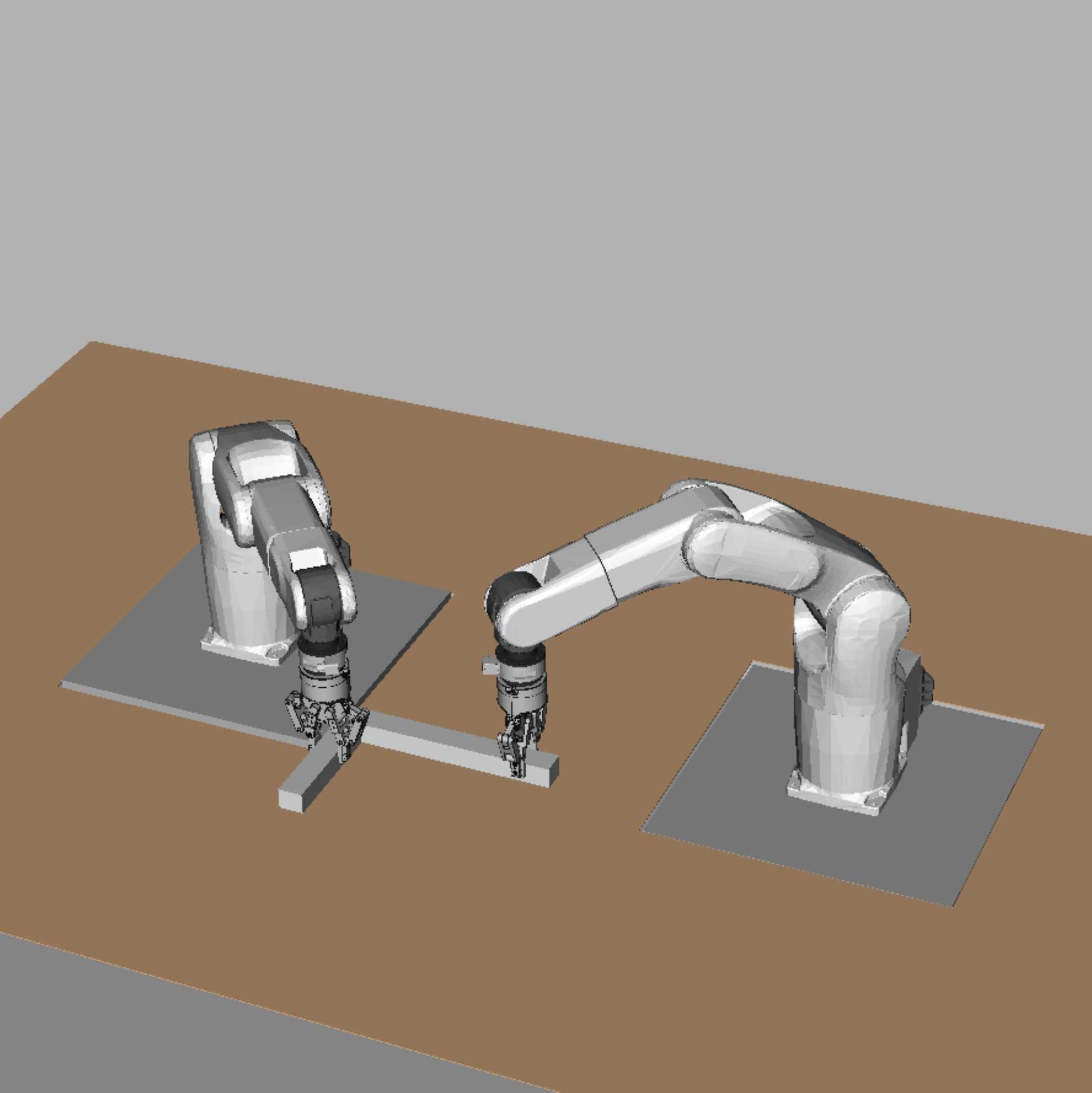}}\hspace{-3pt}
  \subfloat[]{\includegraphics[width=0.12\textwidth]{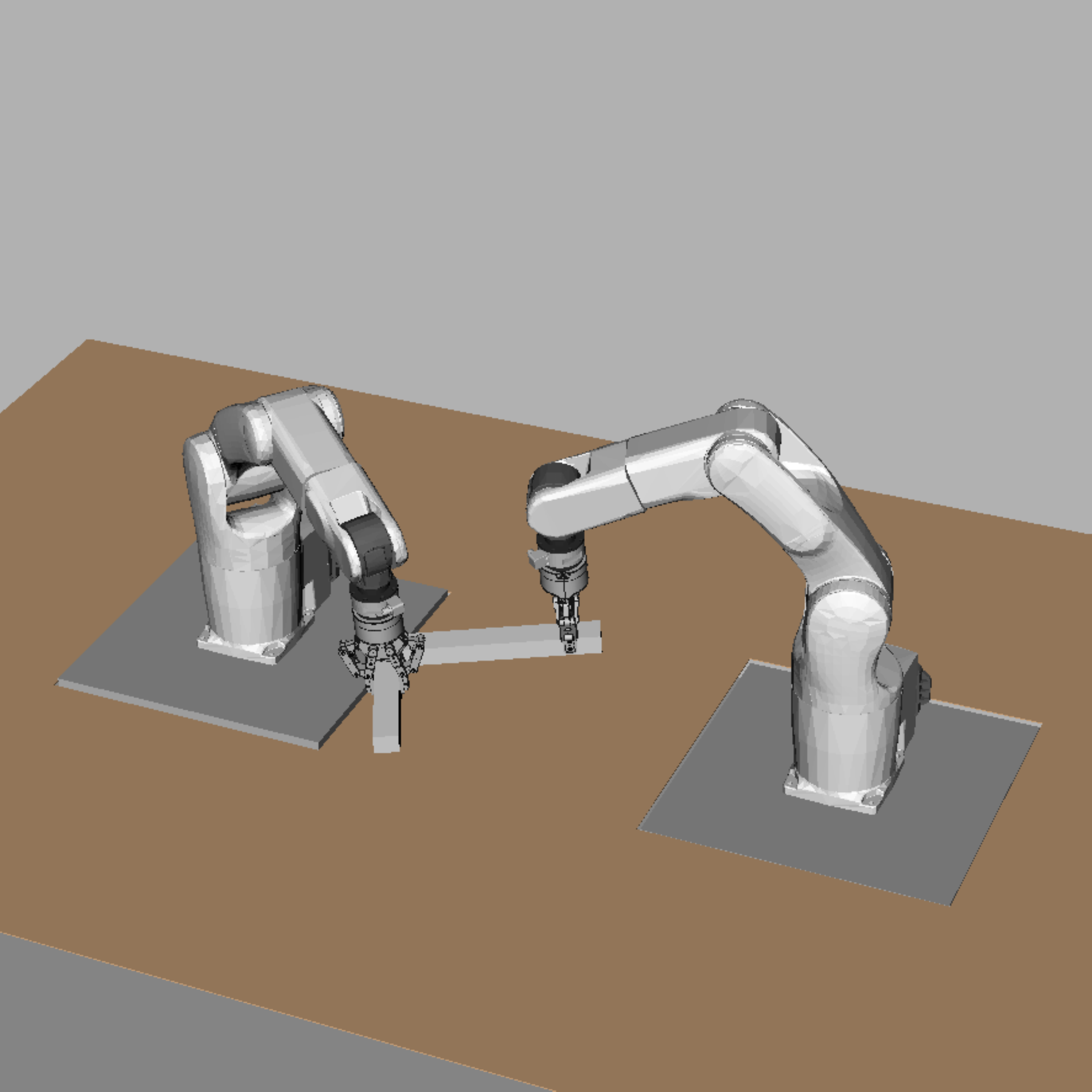}}\hspace{-3pt}
  \subfloat[]{\includegraphics[width=0.12\textwidth]{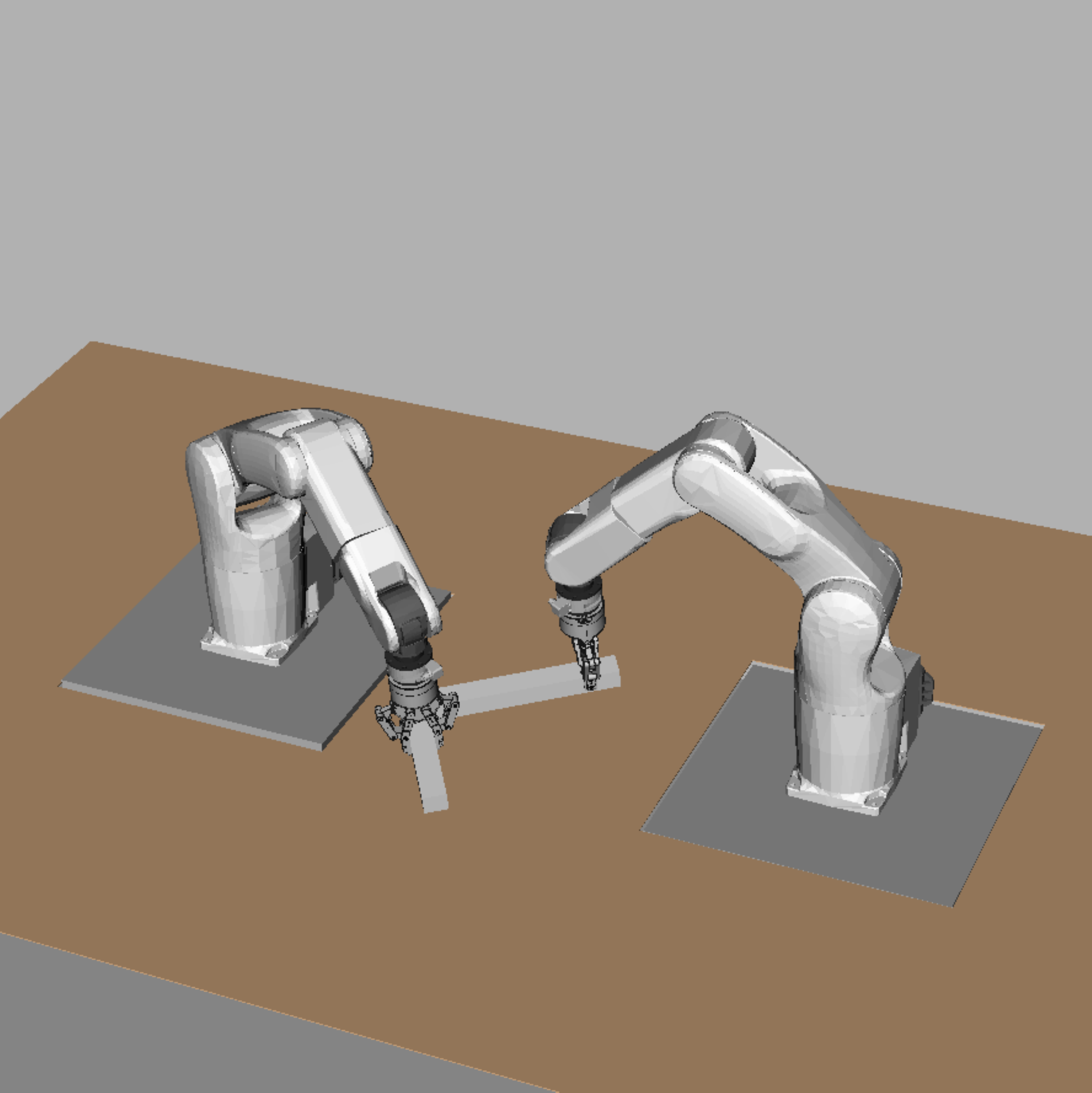}}\hspace{-3pt}
  \subfloat[]{\includegraphics[width=0.12\textwidth]{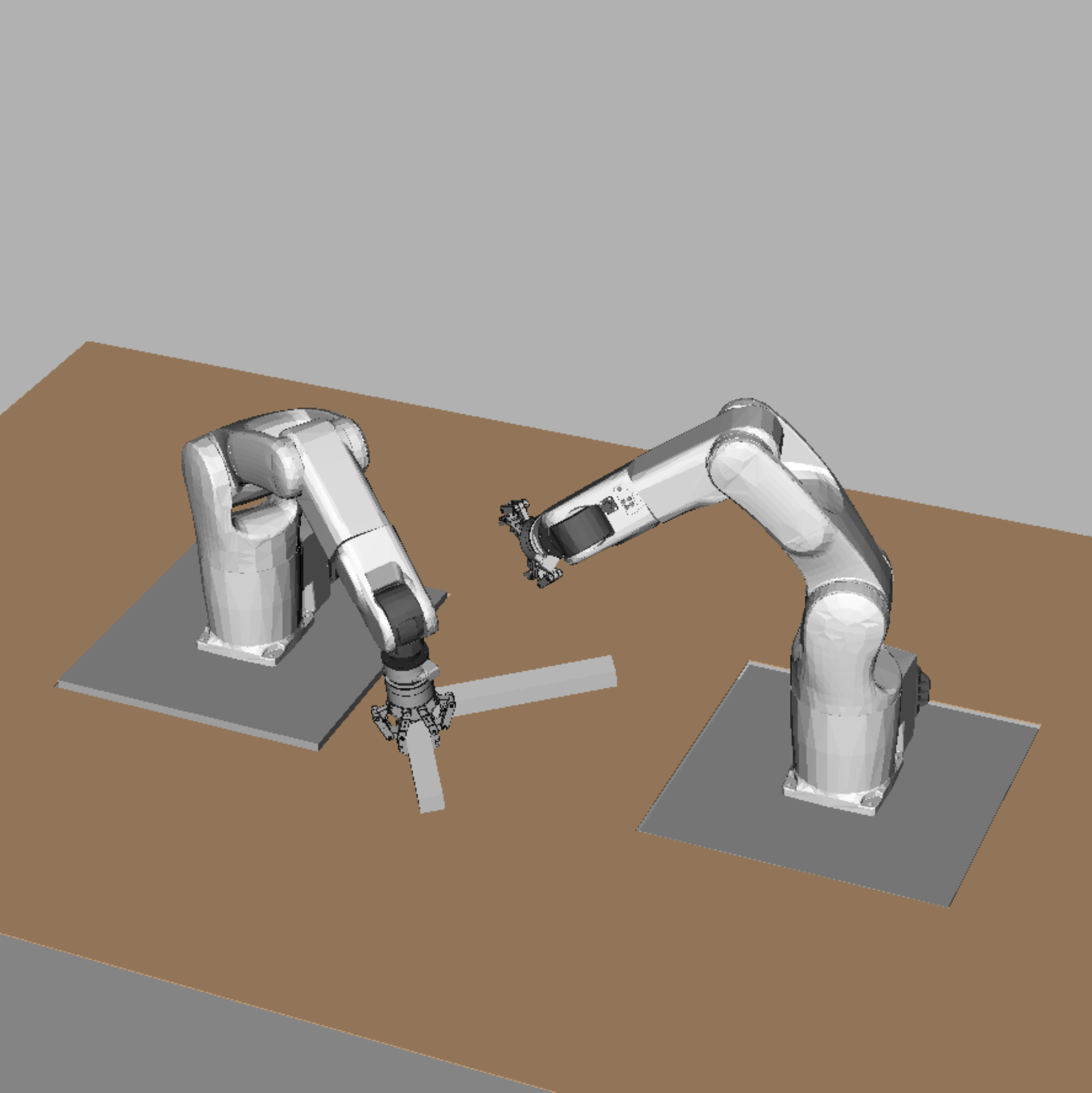}}
  \\\vspace{-8pt}
  \subfloat[]{\includegraphics[width=0.12\textwidth]{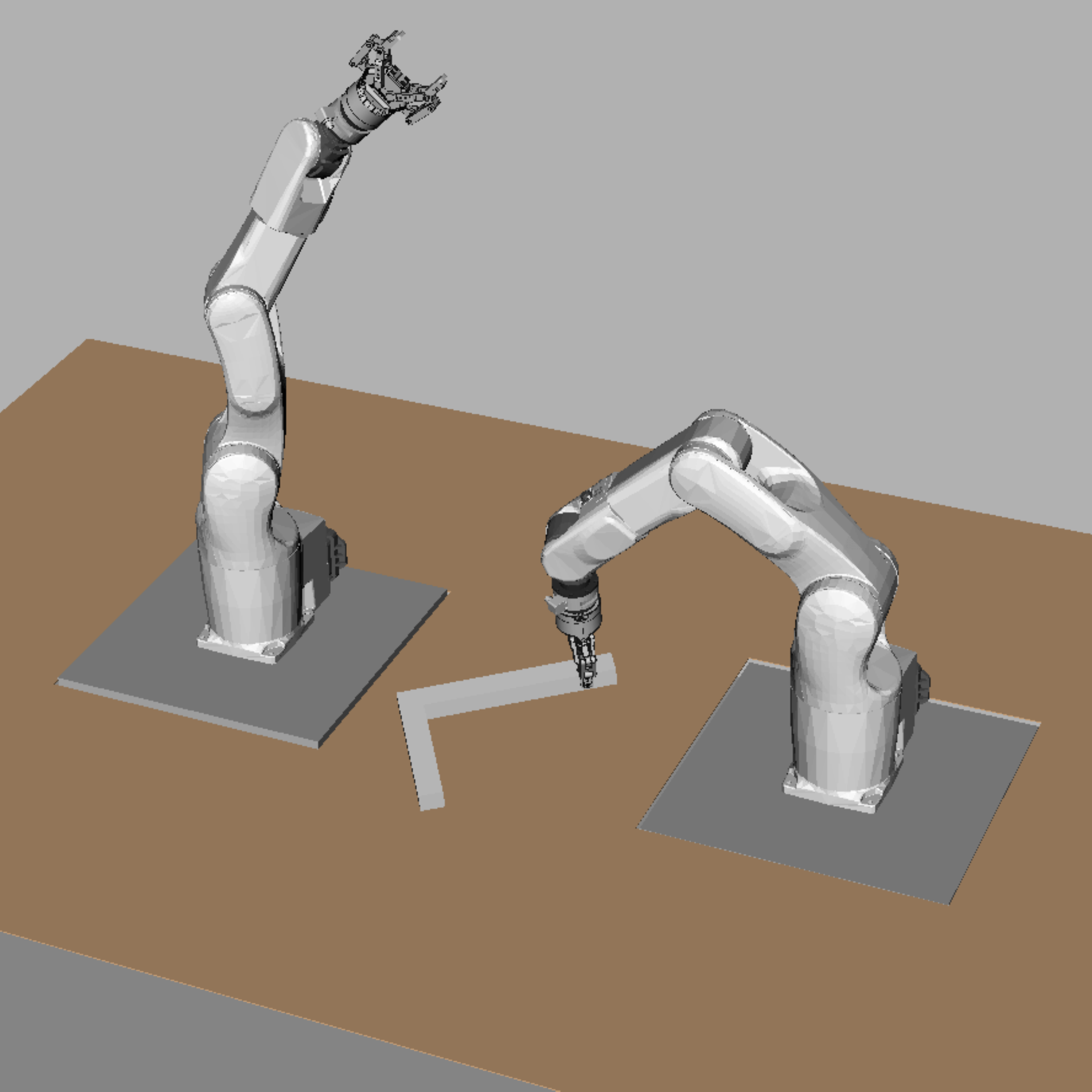}}\hspace{-3pt}
  \subfloat[]{\includegraphics[width=0.12\textwidth]{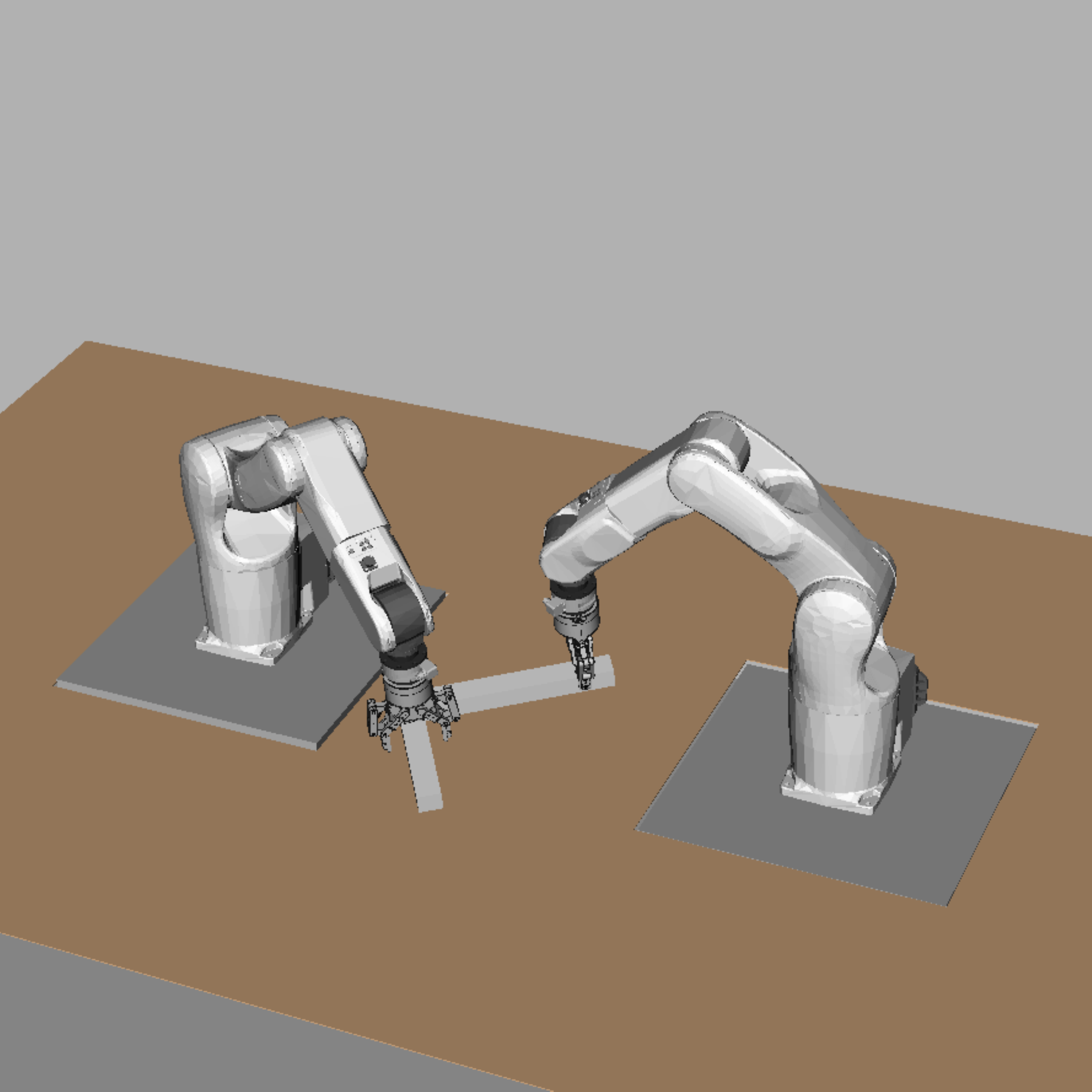}}\hspace{-3pt}
  \subfloat[]{\includegraphics[width=0.12\textwidth]{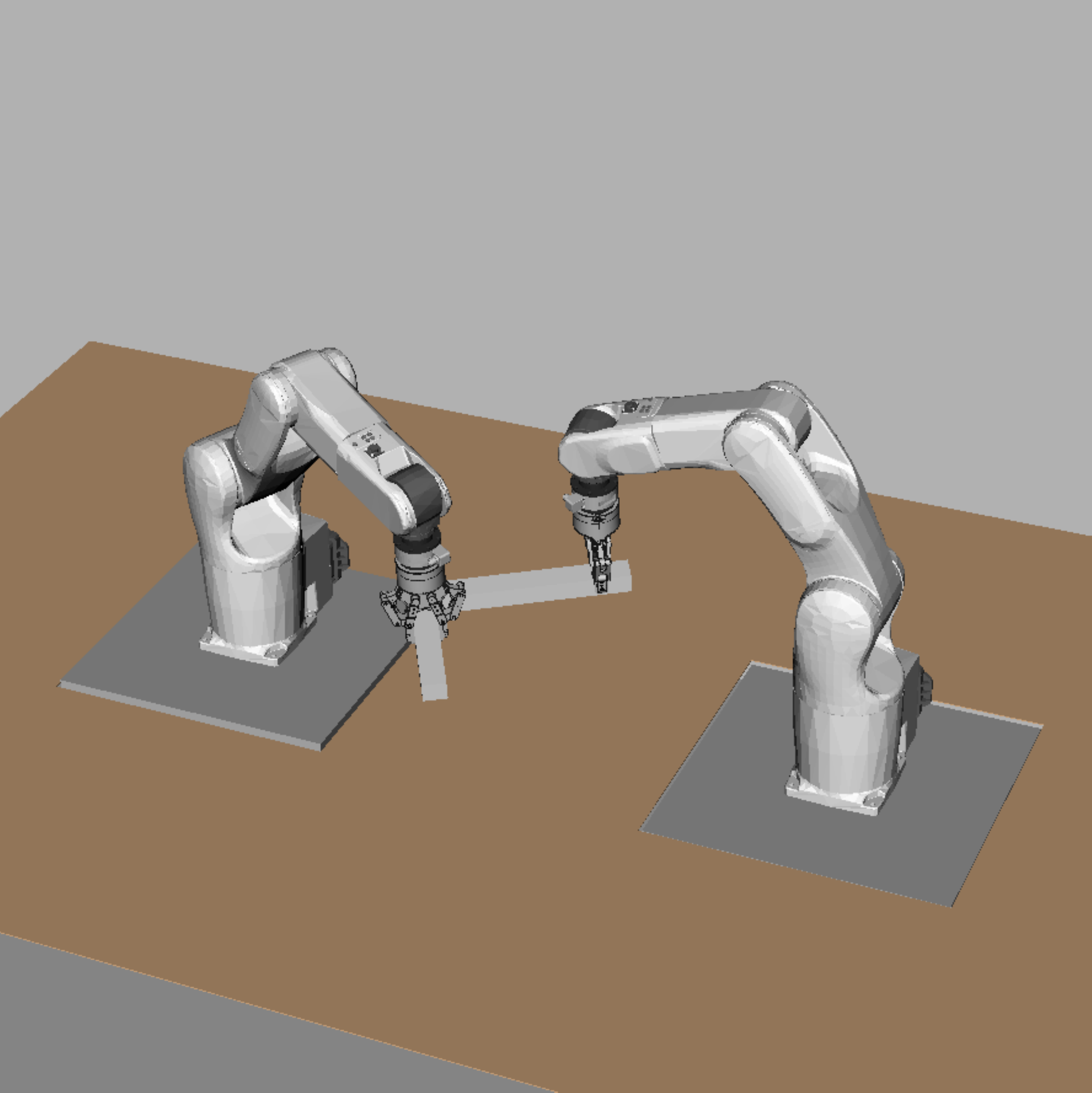}}\hspace{-3pt}
  \subfloat[]{\includegraphics[width=0.12\textwidth]{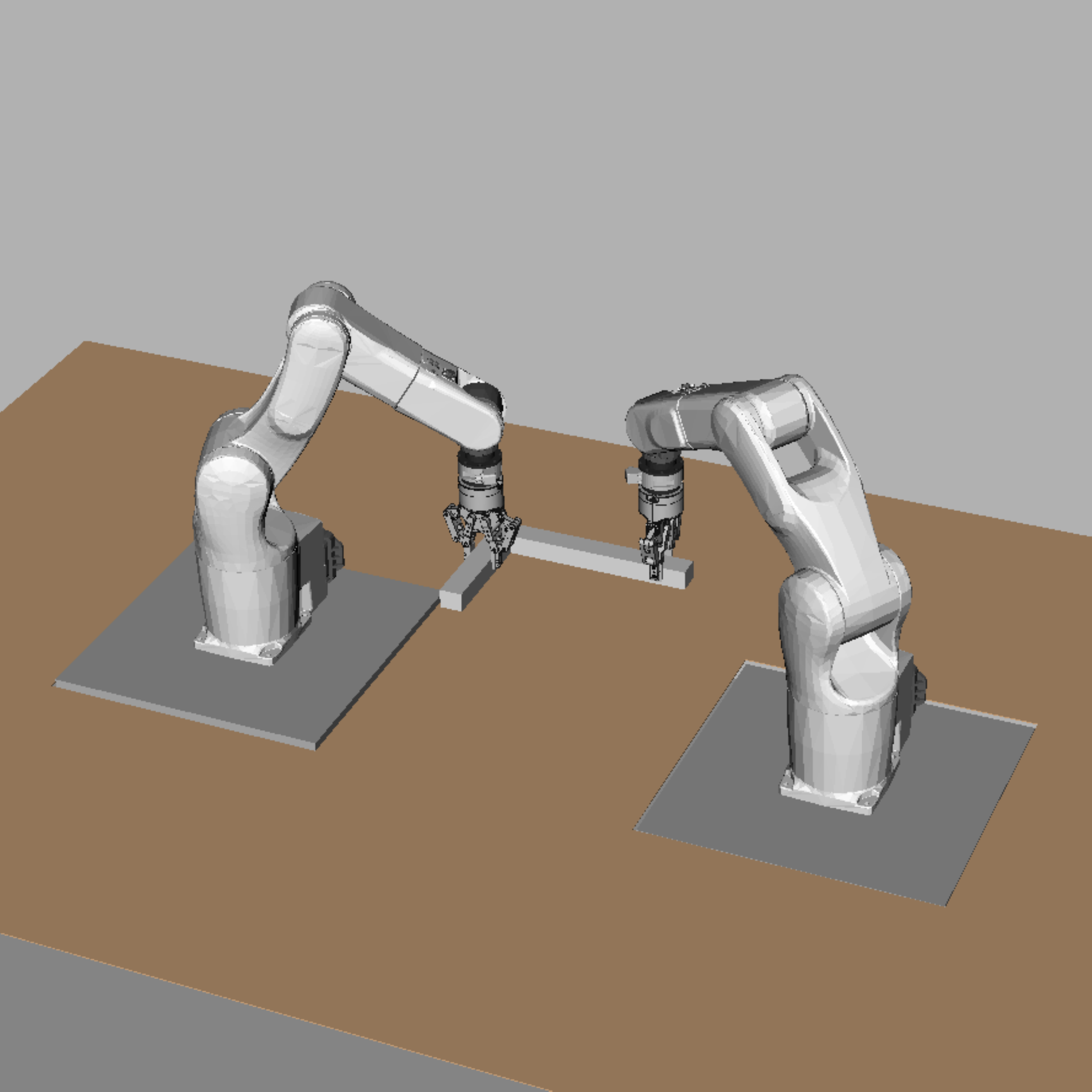}}
  \caption{Snapshots of robots completing Task 1. (a)-(c): Closed-chain
    motion from start to placement configuration,. (d)-(h): IK-switch moves. 
    (f)-(h): Closed-chain
    motion from placement to goal configuration.}
  \label{fig:sim1}
\end{figure}

{\small
Two tasks as shown in Fig. \ref{fig:sim1} and
Fig. \ref{fig:switch}(b, c) are designed to test our planner. Task 1 is a
relatively simple transportation task, where the two robots needs to
move a heavy L-shaped object from a start configuration to a
predesignated goal transformation. In Task 2, the planner is required
to plan a composite path to move and flip a chair frame, which is
a useful operation in common assembly tasks. One goal configuration
$\bms{c}{goal}$ is randomly selected for each tasks (shown in
Fig. \ref{fig:sim1}(h) and Fig.~\ref{fig:switch}(c)). The start and the
goal configuration are verified to be \emph{essentially mutually disconnected},
as sending them to a local closed-chain planner yields no solution in
half an hour.

To ensure the quality of the composite trajectory, the allowed number of
regraspings in Task 1 and Task 2 are set as 1 and 3, respectively. We ran our
planner 50 times for each task. The total planning time, its
decomposition in seconds and number of failures encountered in the second planning
stage were averaged and reported in Table~\ref{tab:table1}.
\vspace{-10pt}

\begin{table}[h!]
  \footnotesize
    \centering
  \caption{Average planning time for each task.\vspace{-8pt}}
  \label{tab:table1}
  \begin{tabular}{p{25pt}<{\centering}p{35pt}<{\centering}p{35pt}<{\centering}p{25pt}<{\centering}p{25pt}<{\centering}p{25pt}<{\centering}}
    \toprule
    &\begin{tabular}{c} \# regrasp\\ limit\end{tabular}  & \# failures & \begin{tabular}{c}global \\ planning\end{tabular} & \begin{tabular}{c}regrasp\\ planning\end{tabular} & total \\ 
    \midrule
    Task 1 & 1 & 4.7 & 0.36 & 2.43 &  2.79 \\
    Task 2 & 3 & 0.02 & 3.66 & 12.15 & 15.81 \\
    \bottomrule
  \end{tabular}
\end{table}
\vspace{-6pt}

In both cases, the planner is able to find a feasible composite path
comprising a series of closed-chain motions and IK-switch moves within a
reasonable amount of time. It can be seen that the IK-switch regrasping planning
occupies a crucial part in the total planning time, since the second stage
involves planning of multiple regrasping trajectories as well as static
equilibrium checkings. Snapshots of the bimanual system executing planned
trajectories (after being smoothened) to complete Task 1 are shown in
Fig. \ref{fig:sim1}. For Task 2, it is implemented on the real hardware
system as explained in Section \ref{sec:control} below (see snapshots in
Fig. \ref{fig:exe}).
}

\subsection{Trajectory Execution and Control}
\label{sec:control}
\begin{figure}
\centering
  \subfloat[]{\includegraphics[width=0.12\textwidth]{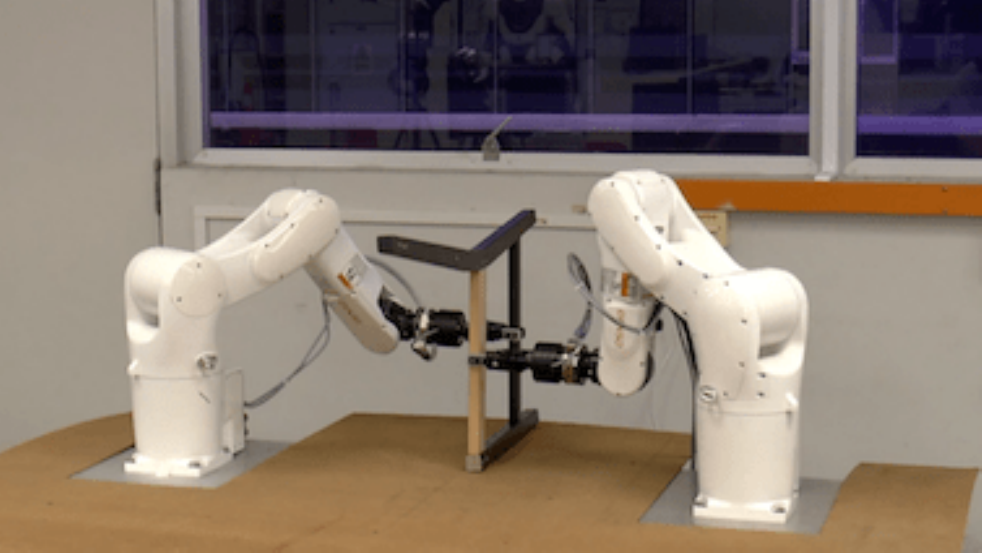}}\hspace{-3pt}
  \subfloat[]{\includegraphics[width=0.12\textwidth]{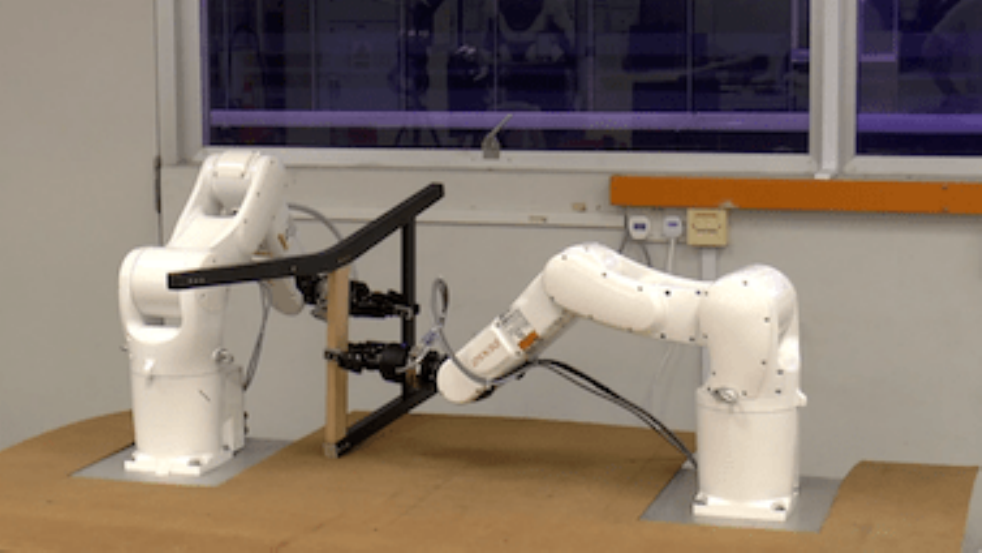}}\hspace{-3pt}
  \subfloat[]{\includegraphics[width=0.12\textwidth]{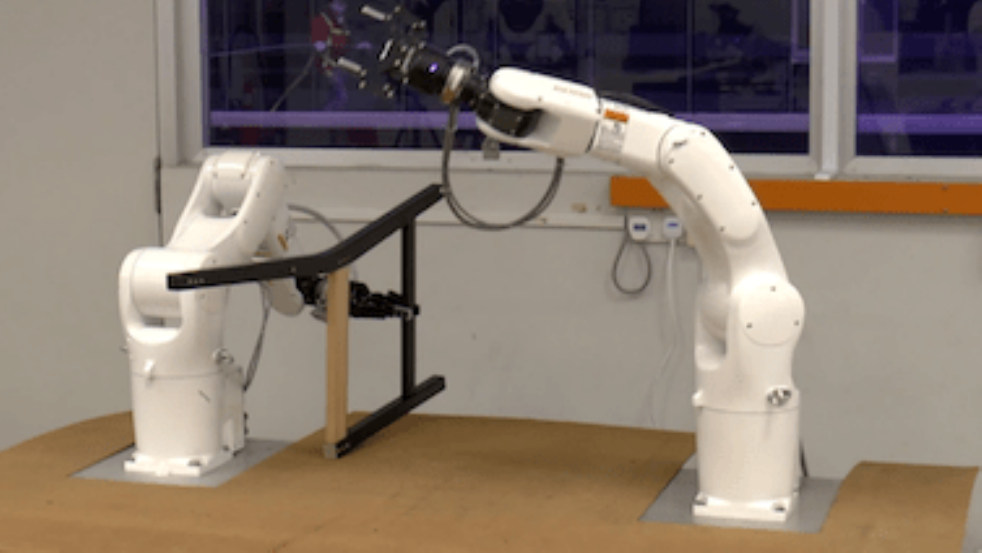}}\hspace{-3pt}
  \subfloat[]{\includegraphics[width=0.12\textwidth]{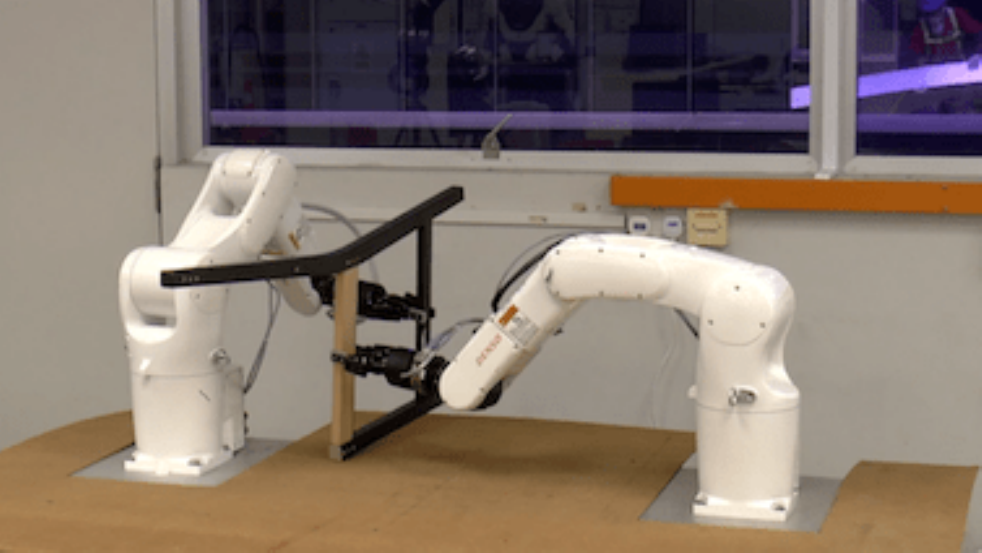}}
  \\\vspace{-8pt}
  \subfloat[]{\includegraphics[width=0.12\textwidth]{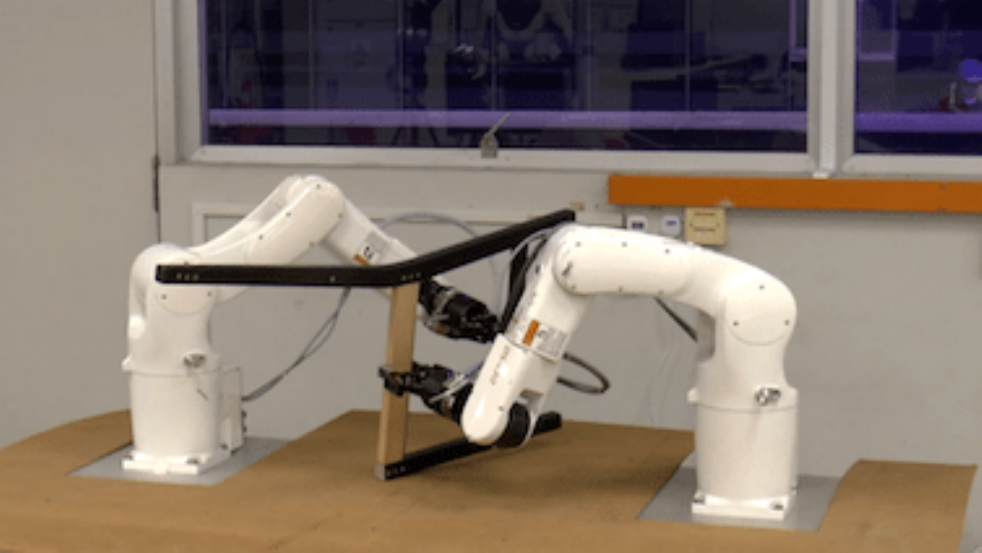}}\hspace{-3pt}
  \subfloat[]{\includegraphics[width=0.12\textwidth]{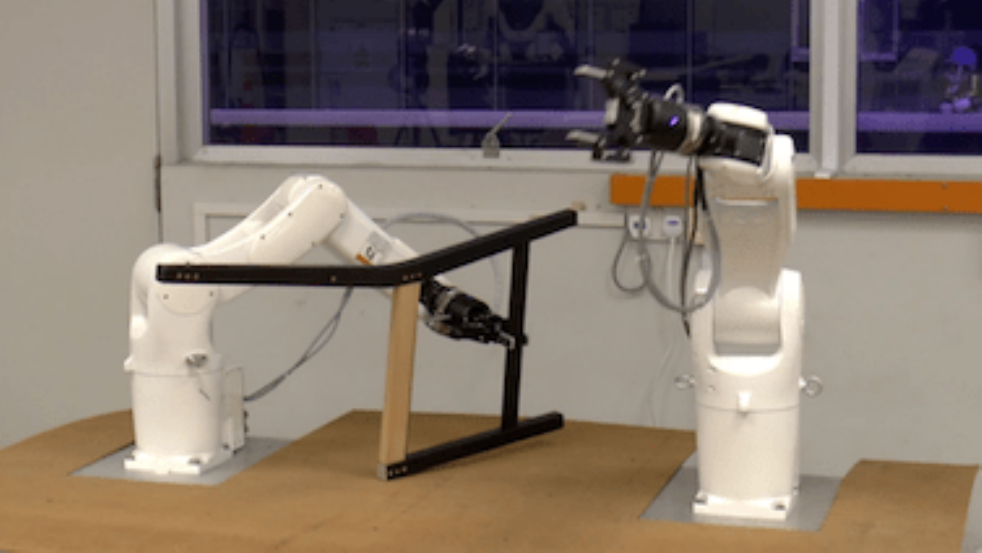}}\hspace{-3pt}
  \subfloat[]{\includegraphics[width=0.12\textwidth]{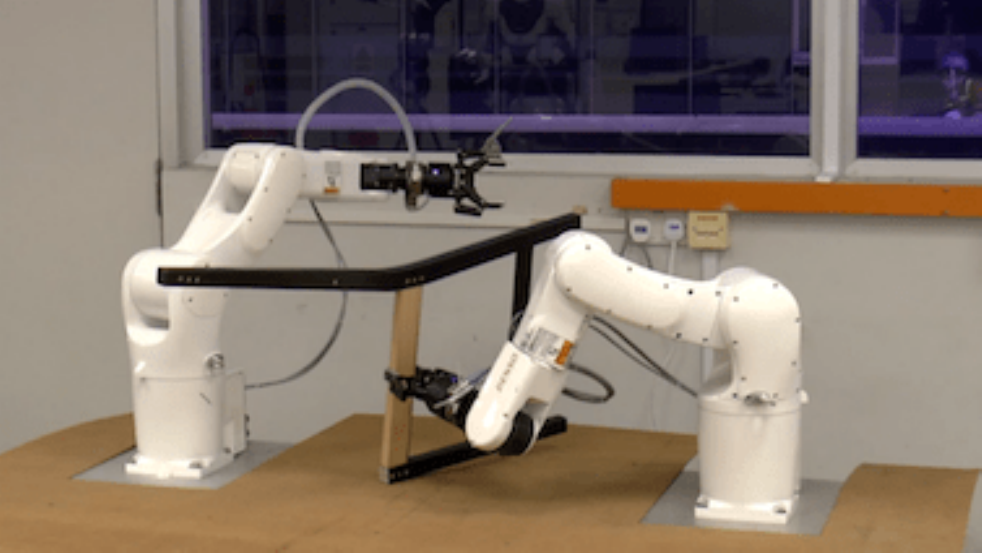}}\hspace{-3pt}
  \subfloat[]{\includegraphics[width=0.12\textwidth]{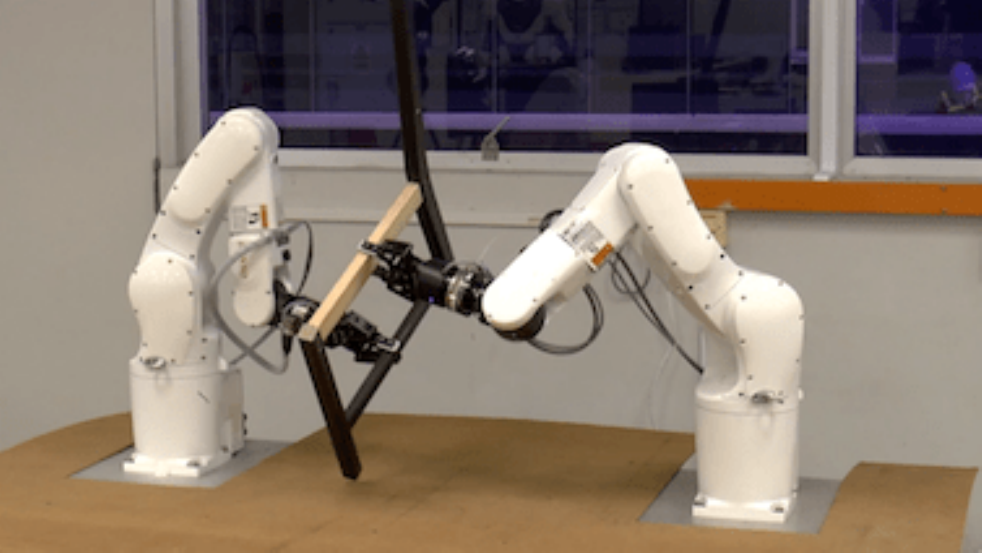}}
  \\\vspace{-8pt}
  \subfloat[]{\includegraphics[width=0.12\textwidth]{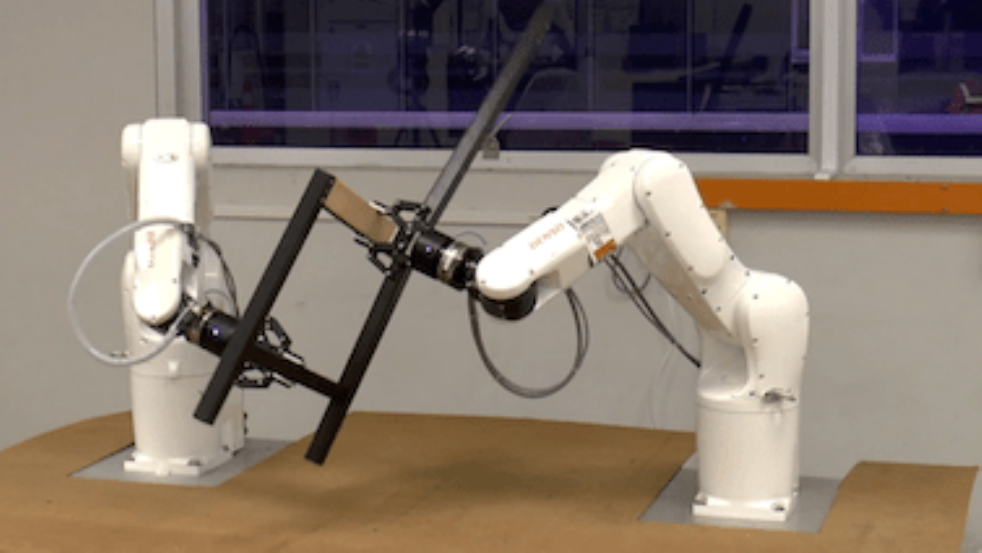}}\hspace{-3pt}
  \subfloat[]{\includegraphics[width=0.12\textwidth]{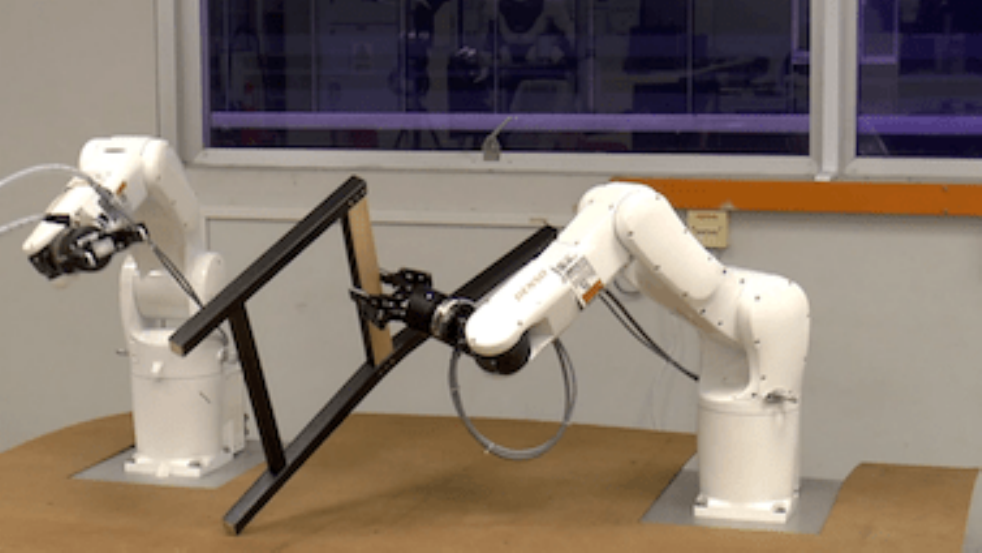}}\hspace{-3pt}
  \subfloat[]{\includegraphics[width=0.12\textwidth]{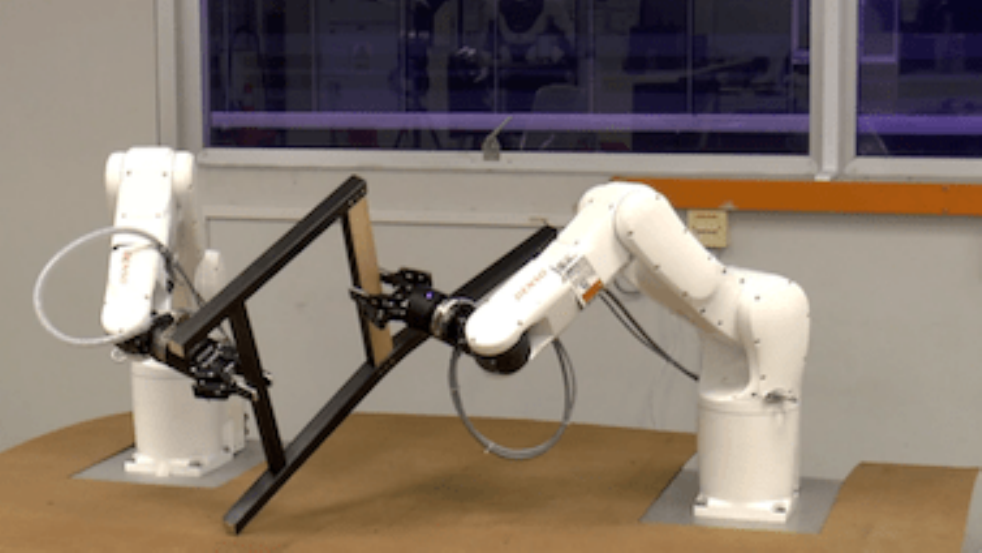}}\hspace{-3pt}
  \subfloat[]{\includegraphics[width=0.12\textwidth]{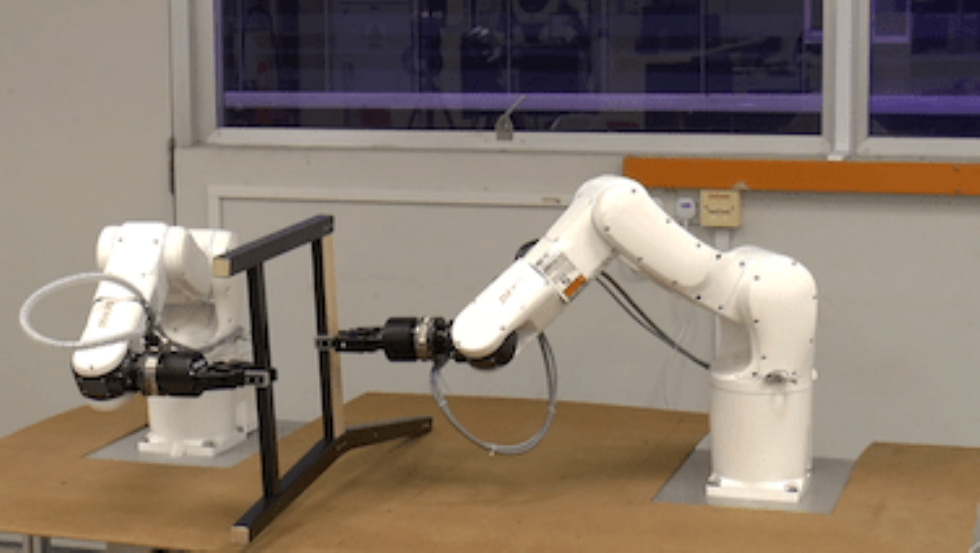}}
  \caption{Snapshots of the bimanual hardware system completing Task 2. (a,
    b, d, e, h, i, k, l): Closed-chain motion. (c, f, g, j): IK-switch moves.}
  \label{fig:exe}
\end{figure}

{\small
A composite path generated by our planner comprises both open-chain
and closed-chain motions. In open-chain motions where each robot
performs IK-switch independently, the robot has no interaction with
the surroundings and thus can be controlled freely via position
control. For closed-chain motions, however, the closed-chain
constraint needs to be satisfied at every time instant and each robot
interacts with others through the object they are grasping. Using pure
position control, small discrepancies between the simulation models
and the real environment as well as robot precision errors can cause
serious damage to the manipulated object. Therefore, we introduce
compliance into our control method by using position-based force
control~\cite{seraji1994adaptive,su2016framework}.

The robots were controlled in a leader-follower fashion. The leader
robot was solely position controlled while the follower robot executed
motion with compliance added. In a discrete-time form with a system
sampling time of $\Delta t$, at each time instant $k$, the target joint
value $\bm{q}[k]$ of the follower is a summation of the theoretical
value $\bm{q}_{t}[k]$ (given by the planner) and a compliance margin
$\bm{q}_{c}[k]$.

In particular, we read the feedback $\bm{f}_{r}$ from the F/T sensor
attached to the follower's wrist and produce a perturbation $\bm{x}_{f}$,
given by $\bm{x}_{f}[k] = \bm{k}_{p}\bm{f}_{e}[k] + \bm{k}_{v}\dot{\bm{f}_{e}}[k]$, where $\bm{f}_{e} = \bm{f}_{i} - \bm{f}_{r}$ is the difference between the
ideal contact force $\bm{f}_{i}$ and the real force $\bm{f}_{r}$, and
$\dot{\bm{f}_{e}}[k] = (\bm{f}_{e}[k] - \bm{f}_{e}[k-1])/\Delta t$.  We
then compute the required perturbation in $\mcs{C}{robot}$ as
$
\bm{q}_{f}[k] = \bm{J}^{-1}\bm{x}_{f}[k]
$
where $\bm{J}$ is the follower's Jacobian matrix. The compliance margin
$\bm{q}_{c}[k]$ is then given by
\begin{equation}
\bm{q}_{c}[0] = \bm{0}, \bm{q}_{c}[k] = \bm{q}_{c}[k-1] + \bm{q}_{f}[k]
\end{equation}
And finally, the target joint value at time instant $k$ of the follower is
set to be
$\bm{q}[k] = \bm{q}_{t}[k] + \bm{q}_{c}[k]$.

This compliant control approach is able to substantially reduce the stress
introduced by modeling errors and the robot's hardware imprecision,
resulting in successful executions of the planned composite trajectory. We
implemented the trajectory planned for Task 2 (as explained in Section
\ref{sec:task}) on our hardware system with this compliant control
strategy. The robots in our bimanual setup were able to execute the
trajectory smoothly and successfully completed the desired task. Snapshots of the
system performing the task are shown in Fig. \ref{fig:exe}. The complete demonstration,
together with the planned trajectory for both task 1 and 2 in simulation,
can be found in the accompanying video.
}

\section{Conclusion}
\label{sec:conclusion}
{
This paper presents a path planner for multi-arm systems manipulating
a large/heavy object. Such a system is constrained by closed-chain
constraints and planning without breaking the kinematic chain may be
practically ineffective or even impossible. We proposed a planning
algorithm which effectively deals with such issues. The algorithm
utilizes \mbox{\emph{regraspings}} to bridge \emph{essentially mutually disconnected} space components together and hence allows us to solve queries that are
otherwise considered as no-solution.

The planner plans a global path first. Then
it plans regrasping moves termed as ``IK-switch'' to complete the global
path. When planning for IK-switch, we resort to the environment in vicinity
to provide support for the object to maintain contact stability. We also
presented an efficient data structure which reorganizes itself to reduce
information loss when certain vertex has to be discarded from a planning
tree. Finally, we presented a compliant control method for closed-chain
trajectory execution. We illustrated effectiveness of our planner via two difficult bimanual
manipulation tasks. With the control method proposed, we also
successfully executed closed-chain trajectories on real hardware.

Our method still contains a number of limitations. First of all, the
planner uses a single pre-determined grasping pose throughout the
planning process. This has the advantage of not requiring the prior
construction of grasp classes~\cite{LP15ral}. However, the possibility
of choosing other grasping poses after regrasping would definitely
increase the flexibility of the planner. Secondly, when exploring
possible placement configurations in current demonstrations, we assume
a planar workspace, while the ability to handle more complex
environment is necessary for practical problems. Our future work will
address these issues so that the planner can deal with more complex
tasks.}

\bibliographystyle{IEEEtran}
{\tiny
\bibliography{references}
}

\end{document}